\documentclass{article}

\usepackage{microtype}
\usepackage{graphicx}

\usepackage{booktabs} 
\usepackage[T1]{fontenc}
\usepackage{multirow} 
\usepackage{hyperref}

\usepackage[accepted]{icml2026}

\usepackage{amsmath}
\usepackage{amssymb}
\usepackage{mathtools}
\usepackage{amsthm}
\usepackage{subcaption}
\usepackage{wrapfig}
\usepackage[capitalize,noabbrev]{cleveref}
\usepackage{caption}  
\usepackage{graphicx} 
\usepackage{booktabs} 
\usepackage{multirow} 
\theoremstyle{plain}
\newtheorem{theorem}{Theorem}[section]

\newtheorem{lemma}[theorem]{Lemma}

\theoremstyle{definition}

\newtheorem{assumption}[theorem]{Assumption}
\theoremstyle{remark}
\newtheorem{remark}[theorem]{Remark}

\usepackage[textsize=tiny]{todonotes}

\icmltitlerunning{}

\begin{document}

\twocolumn[
\icmltitle{AWPO: Enhancing Tool-Use of Large Language Models through Adaptive Integration of Reasoning Rewards}




\icmlsetsymbol{equal}{*}
\icmlsetsymbol{corr}{$\dagger$}  
\begin{icmlauthorlist}
\icmlauthor{Zihan Lin}{equal,ia,mt}
\icmlauthor{Xiaohan Wang}{equal,mt}
\icmlauthor{Hexiong Yang}{ia}
\icmlauthor{Jiajun Chai}{mt}
\icmlauthor{Jie Cao}{ia}
\icmlauthor{Guojun Yin}{corr,mt}
\icmlauthor{Wei Lin}{mt}
\icmlauthor{Ran He}{corr,ia}
\end{icmlauthorlist}

\icmlaffiliation{ia}{Institute of Automation, Chinese Academy of Sciences}
\icmlaffiliation{mt}{Meituan}

\icmlcorrespondingauthor{Guojun Yin}{yinguojun02@meituan.com}
\icmlcorrespondingauthor{Ran He}{ran.he@ia.ac.cn}


\vskip 0.3in
]



\printAffiliationsAndNotice{} 

\begin{abstract}
While Reinforcement Learning (RL) shows promise in training tool-use Large Language Models (LLMs) using verifiable outcome rewards, existing methods largely overlook the potential of reasoning rewards based on chain-of-thought quality for better tool utilization. Furthermore, naïvely combining reasoning and outcome rewards may yield suboptimal performance or conflict with the primary optimization objective. To address this, we propose Advantage-Weighted Policy Optimization (AWPO), a principled RL framework that adaptively integrates reasoning rewards into advantage estimation to improve tool-use performance. AWPO incorporates variance-aware gating and difficulty-aware weighting to adaptively modulate advantages from reasoning signals based on group-relative statistics, alongside a tailored clipping mechanism for stable optimization. Extensive experiments demonstrate that AWPO achieves state-of-the-art performance across standard tool-use benchmarks, significantly outperforming strong baselines and leading closed-source models in challenging multi-turn scenarios. Notably, with exceptional parameter efficiency, our 4B model surpasses Grok-4 by 16.0\% in multi-turn accuracy while preserving generalization capability on the out-of-distribution MMLU-Pro benchmark.
\end{abstract}

\section{Introduction} 


\label{intro}

Recently, tool-use has emerged as a critical capability for extending the functional scope of Large Language Models (LLMs) beyond their inherent parametric knowledge \citep{qin2023toolllm}. Regarding post-training for tool-use LLMs, Supervised Fine-Tuning (SFT) is often constrained by overfitting to demonstration trajectories \citep{fu2025srft}. In contrast, Reinforcement Learning (RL) promotes exploration, thereby mitigating these constraints and enhancing generalization \citep{yue2025promoting}. Consequently, research focus has increasingly shifted toward developing advanced reward systems and RL algorithms specialized for tool-use scenarios \citep{zeng2024agenttuning}.

\begin{figure*}[t]
    \centering
    \includegraphics[width=0.99\linewidth]{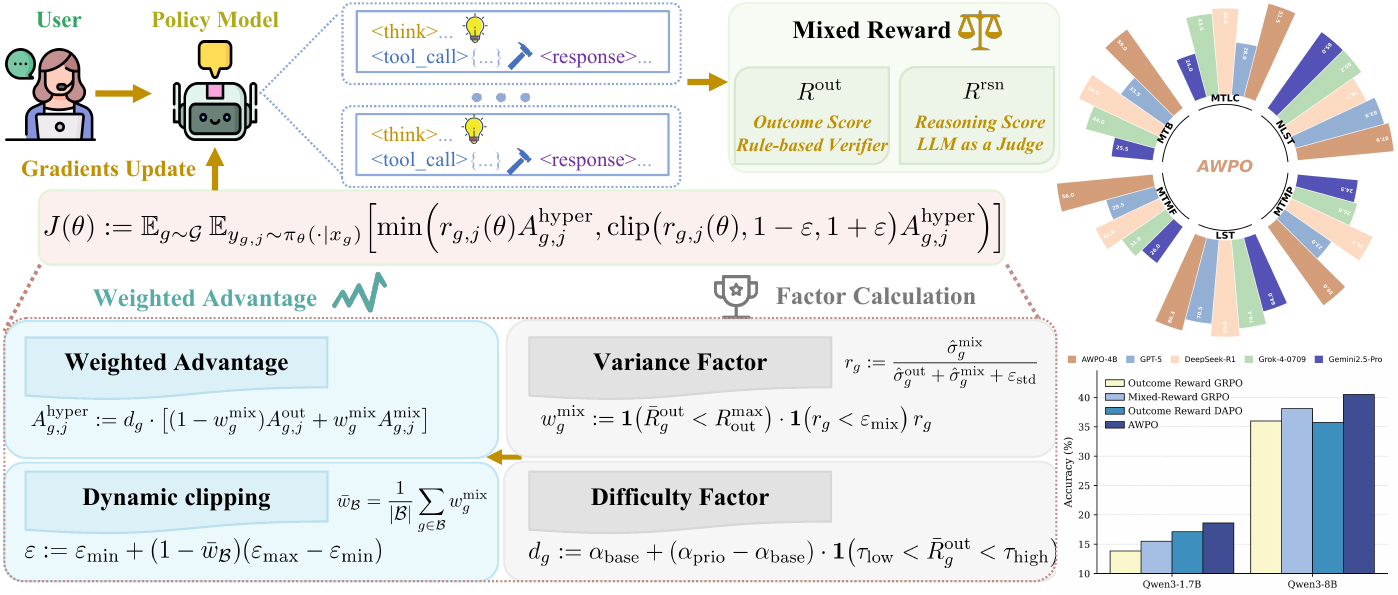}
    \caption{Overview of the AWPO framework and performance. The left panel illustrates the variance-aware gating, difficulty-aware weighting, and dynamic clipping mechanisms. The right panel reports the evaluation results on BFCL. NLST: None-Live Single Turn, MTLC: Multi-Turn Long Context, MTB: Base, MTMF: w/ Missing Functions, LST: Live Single Turn, MTMP: w/ Missing Parameters.}
    \label{fig:main}
    \vspace{-1.5em}
\end{figure*}

However, most existing RL approaches neglect the utilization of reasoning rewards based on chain-of-thought quality to enhance the tool-use capabilities of LLMs. Given that externalized reasoning processes significantly improve accuracy and reliability in complex tasks \citep{wei2022chain}, we hypothesize that providing feedback on reasoning quality, such as the logical coherence of a plan or the appropriateness of tool selection, can substantially improve performance. A naive strategy involves directly combining reasoning reward signals from judge models with outcome rewards. Yet, such simple integration may compete with the primary end-to-end optimization objective, yielding limited and unstable gains. As illustrated in the lower-right panel of Figure~\ref{fig:main} and~\ref{fig:3}, simply mixing reasoning rewards (``Mixed Reward GRPO'') yields inconsistent gains over standard Group-Relative Policy Optimization (GRPO) and is occasionally outperformed by baselines such as DAPO \citep{yu2025dapo}. Based on theoretical analysis, we attribute this limited improvement to optimization instability caused by conflicts between reasoning and outcome rewards. This raises a critical question: \textbf{How can we effectively leverage reasoning rewards to enhance LLM tool-use capabilities without compromising the primary optimization objective?}

To address this, we analyze the underlying causes of reward conflict and redesign the integration mechanism within the RL framework. As illustrated in Figure~\ref{fig:main}, the proposed Advantage-Weighted Policy Optimization (AWPO) adaptively modulates the influence of reasoning rewards based on group-relative statistics. Our approach is grounded in a key insight: AWPO selectively incorporates reasoning reward signals when the variance induced by the outcome reward is statistically insignificant. This ensures that reasoning guidance is introduced judiciously, mitigating optimization conflicts while enhancing reasoning fidelity. Our main contributions are summarized as follows:

\begin{figure*}[h]
    \centering
    \includegraphics[width=1.00\linewidth]{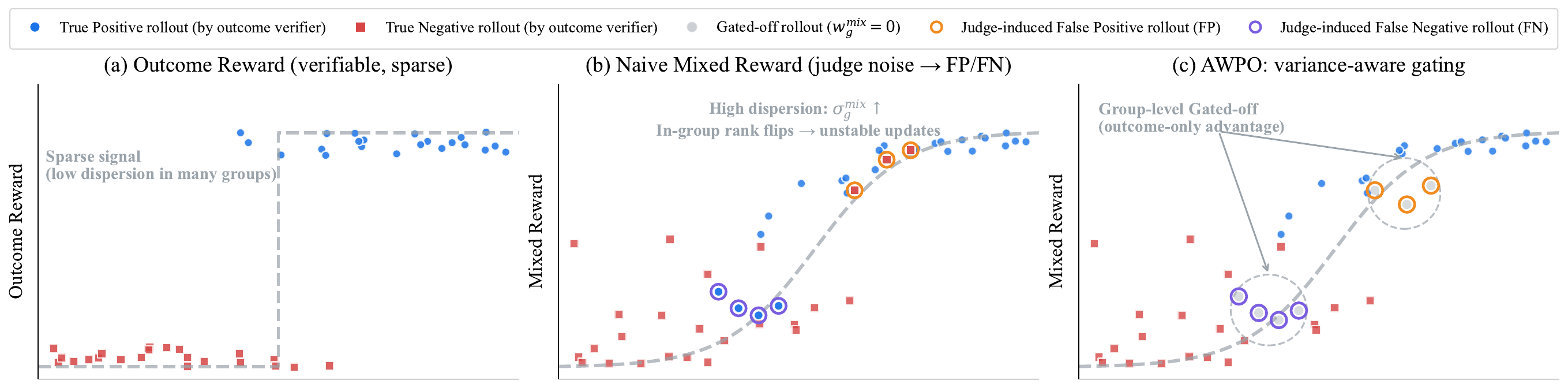}
    \caption{Comparative illustration of reward shaping mechanisms in RL fine-tuning. Verifiable outcome rewards (a) provide reliable but sparse supervision, resulting in low dispersion that limits the optimization potential in many sample groups. Naïve integration of reasoning rewards (b) introduces significant variance and judge-induced noise (e.g., false positives and negatives), which disrupts the within-group ranking and destabilizes training. AWPO (c) mitigates these issues through variance-aware gating, which dynamically filters out high-dispersion groups to ensure that reasoning signals are incorporated only when they offer stable and high-fidelity guidance.}
    \label{fig:111}
    \vspace{-1.5em}
\end{figure*}

1) We establish a theoretical framework for combining reasoning and outcome rewards within a GRPO framework. By deriving an upper bound on the expected policy improvement, we demonstrate that the potential for improvement is governed by a composite signal term, which depends on the alignment of the advantage function with the policy gradient and its variance. This analysis informs the design of the approach, which dynamically adjusts advantage weights to maximize the policy improvement in RL.

2) Building on our theoretical insights, we introduce three novel strategies to realize effective reward integration. First, the variance-aware gating mechanism scales the influence of reasoning rewards based on their discriminative variance relative to outcome rewards within a rollout group. Then, difficulty-aware weighting prioritizes learning from medium-difficulty prompts, where the potential improvement is maximized. To ensure updating stability during backpropagation, we further propose a dynamic clipping strategy that tightens the clipping range.

3) Extensive evaluations on tool-use benchmarks demonstrate that AWPO consistently exceeds strong baselines and the leading closed-source LLMs. In general, AWPO achieves multi-turn accuracy on BFCL by up to $10.50\%$ (a $25.2\%$ relative improvement) and elevates success rates on API-Bank's hardest Level-3 tasks by $15.27\%$ (a $37.7\%$ relative gain) over the baseline RL method. Notably, AWPO enables a 4B base model to surpass closed-source models on tool-use tasks, specifically outperforming Grok-4 by a remarkable margin  $16.00\%$ in multi-turn tool-use accuracy. Additionally, the performance on the out-of-distribution MMLU-Pro benchmark validates the generalization of AWPO by {1.47}{\%} improvements.

\begin{table*}[t]
\caption{Comparison on BFCL benchmark across different models. The column abbreviations stand for: 
           \textbf{OA} (Overall), \textbf{B} (Base), 
           \textbf{MF} (Miss Func), \textbf{MP} (Miss Param), 
           \textbf{LC} (Long Context), \textbf{NL} (Non-Live), 
           and \textbf{L} (Live). Best in \textbf{bold}.}
\label{bfcl1}
  \centering
  \begin{tabular}{@{}lc ccccc ccc@{}}
    \toprule
    \multirow{2}{*}{\textbf{Models}} & \multirow{2}{*}{\textit{Parameter }} & \multicolumn{5}{c}{\textbf{Multi-Turn}} & \multicolumn{3}{c}{\textbf{Single-Turn}} \\
    \cmidrule(lr){3-7} \cmidrule(lr){8-10}
     & & \textit{OA} & \textit{B} & \textit{MF} & \textit{MP} & \textit{LC} & \textit{OA} & \textit{NL} & \textit{L} \\
    \midrule
    GPT-5-2025-08-07 & / & 28.50 & 33.50 & 29.50 & 23.00 & 28.00 & 65.59 & 72.92 & 58.25 \\
    Gemini-2.5-Pro\citep{comanici2025gemini} & / & 25.00 & 25.50 & 26.00 & 24.50 & 24.00 & 74.50 & 85.04 & 63.95 \\
    Grok-4-0709 & / & 36.12 & 44.00 & 31.00 & 26.00 & 43.50 & 79.80 & 85.21 & 74.39 \\
    \midrule
    Moonshotai-Kimi-K2-Inst & 1000B & 41.25 & 51.00 &  {43.00} & 31.00 & 40.00 &  {80.80} & 84.02 & 77.57 \\
    DeepSeek-R1-0528\citep{shao2024deepseekmath} & 671B & 44.50 & 54.50 & 41.00 & 36.50 & 46.00 & 78.22 & 75.73 &  \textbf{80.90} \\
    Qwen3-235B-A22B\citep{yang2025qwen3} & 235B & 40.12 & 49.00 & 41.00 & 29.50 & 41.00 & 82.46 & 87.90 & 77.03 \\
    ToolACE-2-8B\citep{liu2025toolace} & 8B & 37.00 & 47.00 & 31.00 & 28.00 & 42.00 & 82.54 & 87.87 & 77.20 \\
    BitAgent-8B & 8B & 37.75 & 46.50 & 37.50 & 24.00 & 43.00 & 81.71 & 87.33 & 76.09 \\
    watt-tool-8B & 8B & 37.88 & 45.50 & 39.00 & 24.00 & 43.00 & 81.71 & 87.54 & 75.87 \\
        \midrule
    AWPO & 4B & \textbf{52.12} & \textbf{59.00} & \textbf{59.00} & \textbf{39.00} & \textbf{51.50} & \textbf{84.11} & \textbf{87.90} & {80.32} \\
    \bottomrule
  \end{tabular}
\vspace{-1.0em}
\end{table*}

\section{Related Work}
\label{work}
\paragraph{Group Relative Policy Optimization for LLMs}
RL has become a pivotal approach for aligning and improving LLMs, evolving from foundational PPO to more efficient GRPO paradigms ~\citep{ouyang2022training,schulman2017proximal}. GRPO replaces value networks with group-relative baselines for advantage estimation, substantially improving training efficiency and scalability~\citep{shao2024deepseekmath,deepseekai2025deepseekr1incentivizingreasoningcapability}. Subsequent advances have introduced variants to address specific limitations: DAPO stabilizes long-chain reasoning via decoupled advantage clipping~\citep{yu2025dapo}, while Dr.GRPO removes normalizing terms that inadvertently reward verbose outputs~\citep{liu2025drgrpo}. To tackle credit assignment in long-horizon tasks, methods such as GiGPO and IGPO employ structured grouping mechanisms~\citep{feng2025group,zhao2025inpainting}, and BAPO introduces adaptive clipping for stable off-policy updates~\citep{xi2025bapo}. Despite algorithmic progress, GRPO-style methods largely overlook principled integration of reasoning rewards, leaving the benefits of reasoning signals for complex tool use underexploited. Our work bridges this gap by dynamically injecting reasoning signals into advantage estimation, improving both reasoning fidelity and tool-use performance.

\paragraph{Post-Training for Tool-Use}
Enabling LLMs to reliably interact with external tools through multi‑step planning is a core challenge in building agentic systems~\citep{qin2023toolllm,liu2025toolace,zhang2024xlamfamilylargeaction,chen2025toolforgedatasynthesispipeline}. BalanceSFT balances reasoning and function‑call tokens while resampling hard examples ~\citep{hao2025funreasonenhancinglargelanguage}. SwiRL extends this idea by synthesizing multi‑step tool‑use trajectories for step‑wise optimization~\citep{goldie2025syntheticdatageneration}. ToolRL demonstrates that GRPO with decomposed rewards outperforms SFT on tool‑use benchmarks~\citep{qian2025toolrl}, while ResT introduces entropy‑guided gradient updates to refine token‑level decisions~\citep{lin2025rest}. Although these methods advance tool‑use capability, they predominantly rely on outcome rewards and neglect reasoning feedback.

\paragraph{LLM-as-a-Judge for Reward Design}
Automated evaluation of reasoning quality increasingly relies on LLM-as-a-Judge paradigms, where a dedicated LLM scorer rates candidate responses based on structured rubrics~\citep{gu2025surveyllmasajudge}. Early benchmarks such as MT-Bench and AlpacaEval employed GPT-family judges to produce scores that correlate well with human preferences~\citep{bai2024mt,dubois2024length}. Crowdsourced platforms like Chatbot Arena further popularized pairwise judging for model ranking~\citep{chiang2024chatbot,li2024crowdsourced}. Recently, such automated scores have been integrated into RL pipelines to optimize reasoning and code-generation models without human annotation~\citep{lee2023rlaif,lambert2025rewardbench}. To mitigate potential scorer biases, subsequent studies introduced calibration and reweighting techniques~\citep{li2025evaluatingscoringbiasllmasajudge}. Building on these insights, we design a specialized LLM-as-a-Judge module to assess the logical coherence, correctness, and tool-call appropriateness of generated reasoning content.
\section{Method}
\label{method}
\subsection{Theoretical Analysis}
\label{theory}
\paragraph{Preliminaries}
We consider the GRPO clipped objective
\begin{equation}
\label{eq:grpo_obj_recap}
\begin{split}
J(\theta)
&= \mathbb{E}\Bigl[\min\Bigl(
        r_\theta \,\tilde A,\;
        \operatorname{clip}(r_\theta,1-\varepsilon,1+\varepsilon)\,\tilde A
   \Bigr)\Bigr],
\end{split}
\end{equation}
where $r_\theta=\frac{\pi_\theta(a|s)}{\pi_{\theta_{\mathrm{old}}}(a|s)}$, $\tilde A$ is the advantage, and $\varepsilon>0$ is the clip radius.
For each prompt group $g$ we draw $K$ samples and compute within-group normalized advantages
\begin{equation}
A^t_{g,j} \;=\; \frac{R^t_{g,j}-\bar R_g^{\,t}}{\widehat\sigma_g^{\,t}+\epsilon},
\qquad t\in\{\mathrm{out},\mathrm{mix}\},
\end{equation}
with $\widehat\sigma_g^{\,t}=\sqrt{\widehat V_g^{\,t}}$ the within-group reward dispersion.

We optimize verifiable tool-use correctness with $R^{\mathrm{out}}$ and treat $R^{\mathrm{reasoning}}$ as a potentially noisy auxiliary signal. We analyze how mixing affects GRPO’s within-group preference structure and the alignment between outcome-induced and mixed-induced policy gradients.
The mixed reward satisfies $R^{\mathrm{mix}}=R^{\mathrm{out}}+R^{\mathrm{reasoning}}$ by construction, but the {normalized} advantages differ due to distinct within-group scalings.
Define centered rewards $\widetilde R^t_{g,j}\coloneqq R^t_{g,j}-\bar R_g^{\,t}$.

\begin{lemma}[Normalized auxiliary perturbation decomposition]
\label{lem:adv_decomp}
For each group $g$ and sample $j$,
\begin{equation}
\label{eq:adv_decomp}
A^{\mathrm{mix}}_{g,j}
\;=\;
\underbrace{\frac{\widehat\sigma_g^{\mathrm{out}}}{\widehat\sigma_g^{\mathrm{mix}}}\,A^{\mathrm{out}}_{g,j}}_{\text{anchor term}}
\;+\;
\underbrace{\frac{\widetilde R^{\mathrm{reasoning}}_{g,j}}{\widehat\sigma_g^{\mathrm{mix}}+\epsilon}}_{\text{aux perturbation}}
\;+\;
\Delta_{g,j},
\end{equation}
where $\Delta_{g,j}$ is a residual term induced by the stabilization constants in the denominators and disappears in the idealized limit $\epsilon\downarrow 0$.
\end{lemma}

Lemma~\ref{lem:adv_decomp} decomposes $A^{\mathrm{mix}}$ into a rescaled outcome advantage and an auxiliary perturbation induced by the judge reward. When the perturbation dominates, $A^{\mathrm{mix}}$ can change the within-group ranking of samples that underlies group-normalized updates.

Following standard LLM-as-a-Judge practice, we model the judge reward as a biased auxiliary signal that can deviate from, but is not systematically in conflict with, outcome reward \citep{gu2025surveyllmasajudge,li2025evaluatingscoringbiasllmasajudge,ye2025justice}. AWPO controls judge-induced preference bias via ratio-gated mixing and post-saturation annealing.

\begin{assumption}[Order-consistent auxiliary score]
\label{ass:cov_nonneg}
For each group $g$, the auxiliary reward is comonotone with the centered outcome reward in the sense that for any samples $i,j$,
\begin{equation}
\widetilde R^{\mathrm{out}}_{g,i} \ge \widetilde R^{\mathrm{out}}_{g,j}
\;\Rightarrow\;
\mathbb{E}\!\left[\widetilde R^{\mathrm{reasoning}}_{g,i}\right]\ge \mathbb{E}\!\left[\widetilde R^{\mathrm{reasoning}}_{g,j}\right].
\end{equation}
This condition implies $\mathrm{Cov}(\widetilde R^{\mathrm{out}}_g,\widetilde R^{\mathrm{reasoning}}_g)\ge 0$, which is the only property used in the subsequent derivations.
\end{assumption}

\begin{lemma}[Within-group advantage correlation and a dispersion proxy]
\label{lem:corr_ratio}
Under Assumption~\ref{ass:cov_nonneg} and ignoring $\Delta_{g,j}$,
\begin{equation}
\label{eq:corr_ratio_lb}
\mathbb{E}\Big[\mathbb{E}_j\!\left[A^{\mathrm{out}}_{g,j}A^{\mathrm{mix}}_{g,j}\right]\Big]
\;\ge\;
\mathbb{E}\!\left[
\frac{(\widehat\sigma_g^{\mathrm{out}})^2}{(\widehat\sigma_g^{\mathrm{out}}+\epsilon)(\widehat\sigma_g^{\mathrm{mix}}+\epsilon)}
\right].
\end{equation}
Moreover, when $\epsilon$ is small and the covariance contribution in \eqref{eq:exact_corr_decomp} is negligible
(or ignored as a conservative proxy), we have the approximation
\begin{equation}
\label{eq:corr_ratio_app}
\mathbb{E}\Big[\mathbb{E}_j\!\left[A^{\mathrm{out}}_{g,j}A^{\mathrm{mix}}_{g,j}\right]\Big]
\;\approx\;
\mathbb{E}\Big[\frac{\widehat\sigma_g^{\mathrm{out}}}{\widehat\sigma_g^{\mathrm{mix}}}\Big]
\approx
\frac{\sigma_g^{\mathrm{out}}}{\sigma_g^{\mathrm{mix}}}.
\end{equation}
\end{lemma}

By Lemma~\ref{lem:corr_ratio}, the correlation between $\widehat A_g^{\mathrm{mix}}$ and $\widehat A_g^{\mathrm{out}}$ decreases as $\widehat\sigma_g^{\mathrm{mix}}/\widehat\sigma_g^{\mathrm{out}}$ increases, making mixed updates more likely to track auxiliary-induced within-group preferences than outcome-induced ones. We thus use the dispersion ratio as a reliability proxy to control reward mixing.

We next translate this correlation into a gradient-level statement.
Let $Z(s,a)\coloneqq \nabla_\theta \log \pi_\theta(a|s)$ be the score function and
let $g^t_g\coloneqq \mathbb{E}[Z\,A^t\mid g]$ denote the group-wise policy gradient direction induced by $A^t$.

\begin{lemma}[Gradient-alignment proxy via within-group correlation]
\label{lem:grad_alignment}
Assume Assumption~\ref{ass:gram_iso} holds, $\|Z(s,a)\|$ is bounded, and the sampling distribution is fixed within the update.
Then, in expectation over the $K$ samples in group $g$,
\begin{equation}
\label{eq:cos_lower}
\mathbb{E}\!\left[\cos\!\bigl(g^{\mathrm{out}}_g,\,g^{\mathrm{mix}}_g\bigr)\right]
\;\gtrsim_{\delta_g}\;
\mathbb{E}\Big[\mathbb{E}_j\!\left[A^{\mathrm{out}}_{g,j}A^{\mathrm{mix}}_{g,j}\right]\Big],
\end{equation}
where $\gtrsim_{\delta_g}$ denotes a lower bound up to group-dependent constants and a slack term controlled only by the spectral deviation parameter $\delta_g$ (see Appendix~\ref{proof:lemma3}).
\end{lemma}

Lemma~\ref{lem:grad_alignment} links gradient alignment to a within-group correlation proxy.
Combined with Lemma~\ref{lem:corr_ratio}, this motivates using the dispersion ratio $\widehat\sigma_g^{\mathrm{out}}/\widehat\sigma_g^{\mathrm{mix}}$ as an observable alignment-control statistic.

Specifically, AWPO constructs a convex combination of $A^{\mathrm{out}}$ and $A^{\mathrm{mix}}$:
\begin{equation}
A^{\mathrm{hyper}}_{g,j}
\;=\;
d_g\Bigl[(1-w^{\mathrm{mix}}_g)A^{\mathrm{out}}_{g,j}+w^{\mathrm{mix}}_gA^{\mathrm{mix}}_{g,j}\Bigr].
\end{equation}
The following theorem shows that ratio-based gating enforces a uniform lower bound on this proxy.

\begin{theorem}[Safe mixing condition implied by ratio gating]
\label{thm:safe_mixing}
Define the observable ratio
\(
\rho_g\coloneqq \frac{\widehat\sigma^{\mathrm{mix}}_g}{\widehat\sigma^{\mathrm{out}}_g+\widehat\sigma^{\mathrm{mix}}_g+\varepsilon_{\mathrm{std}}}.
\)
Assume $\varepsilon_{\mathrm{std}}$ is negligible ($\epsilon_\text{std}\ll \hat\sigma_\text{mix}$).
If $\rho_g\le \varepsilon_{\mathrm{mix}}<1$, then
\begin{equation}
\label{eq:alignment_threshold}
\frac{\widehat\sigma_g^{\mathrm{out}}}{\widehat\sigma_g^{\mathrm{mix}}}
\;\ge\;
\frac{1-\varepsilon_{\mathrm{mix}}}{\varepsilon_{\mathrm{mix}}}
\;\eqqcolon\;
\kappa_{\min}.
\end{equation}
Consequently, combining Lemma~\ref{lem:corr_ratio} with Lemma~\ref{lem:grad_alignment} yields the alignment-proxy floor
\begin{equation}
\label{eq:alignment_proxy_floor}
\mathbb{E}\!\left[\cos\!\bigl(g^{\mathrm{out}}_g,\,g^{\mathrm{mix}}_g\bigr)\right]
\;\gtrsim_{\delta_g}\;
\mathbb{E}\Big[\mathbb{E}_j\!\left[A^{\mathrm{out}}_{g,j}A^{\mathrm{mix}}_{g,j}\right]\Big]
\;\gtrsim\;
\kappa_{\min},
\end{equation}
where the last relation uses the dispersion-only proxy induced by Lemma~\ref{lem:corr_ratio} (ignoring $\Delta_{g,j}$ and taking $\epsilon$ small).
Therefore, ratio-based gating $r_g<\varepsilon_{\mathrm{mix}}$ enforces a uniform lower bound on the alignment proxy, preventing mixed updates from becoming arbitrarily misdirected within any group.
\end{theorem}

Theorem~\ref{thm:safe_mixing} formalizes the role of Eq.~\ref{eq:mix_weight}: the gating rule depends only on observable within-group dispersions yet yields a conservative alignment proxy floor.
AWPO further multiplies this gate with $\mathbf{1}(\bar R_g^{\mathrm{out}}<R_{\mathrm{out}}^{\max})$.
This saturation gate is consistent with treating $R^{\mathrm{reasoning}}$ as auxiliary: once verifiable correctness saturates on a group, continuing to optimize the judge reward can introduce unnecessary bias toward judge-preferred rationales.

\subsection{Advantage Weighted Policy Optimization}

Section~\ref{theory} shows that reasoning rewards provide useful auxiliary feedback but can also perturb the within-group preference structure that GRPO relies on: although $R^{\mathrm{mix}}=R^{\mathrm{out}}+R^{\mathrm{reasoning}}$, the corresponding normalized advantages can deviate from the outcome-induced ordering when the auxiliary perturbation dominates (Lemma~\ref{lem:adv_decomp}).
Moreover, the dispersion ratio $\widehat\sigma_g^{\mathrm{out}}/\widehat\sigma_g^{\mathrm{mix}}$ yields an observable proxy of mixed-to-outcome alignment (Lemmas~\ref{lem:corr_ratio}--\ref{lem:grad_alignment}), and ratio-based gating enforces a conservative alignment-proxy floor (Theorem~\ref{thm:safe_mixing}).
AWPO therefore injects judge feedback only when it is conservatively proxy-aligned, anneals it once verifiable correctness saturates, prioritizes intermediate-difficulty groups through $d_g$, and tightens the GRPO clip radius as auxiliary reliance increases to control update variance.

\paragraph{Mixed Reward}
We compute the outcome reward $R_j^{\mathrm{out}}$ using deterministic rules \citep{qian2025toolrl} that verify output structure and tool-call correctness.
We decompose $R_j^{\mathrm{out}}$ into a format score and an execution score:
\begin{equation}
{R}^{\text{out}}_j
=  {S}_{j}^{\text{format}}
  +  {S}_{j}^{\text{exec}}.
\end{equation}

The format score uses exact match:
\begin{equation}
{S}_{j,\mathrm{format}} =
\begin{cases}
1, & \text{if the structure is valid},\\
0, & \text{otherwise}.
\end{cases}
\end{equation}

For execution correctness, let $T$ be the ground-truth tool graph and $P$ the prediction. We assess tool names via Jaccard similarity,
\begin{equation}
r_{\mathrm{name}}=\frac{|N_T\cap N_P|}{|N_T\cup N_P|}\in[0,1],
\end{equation}
where $N_T,N_P$ are the ground-truth and predicted tool-name sets; parameter names, aggregated over tools,
\begin{equation}
r_{\mathrm{para}}=\sum_{T_i\in T}\frac{|P_T(T_i)\cap P_P(T_i)|}{|P_T(T_i)\cup P_P(T_i)|},
\end{equation}
and parameter values via exact match,
\begin{equation}
r_{\mathrm{value}}
=\sum_{T_i\in T}\ \sum_{v\in \mathbf{v}(T_i)} \mathbf{1}\big[P_T(T_i)[v]=P_P(T_i)[v]\big].
\end{equation}
The execution score is the normalized sum
\begin{equation}
{S}_{j,\mathrm{exec}}
=\frac{r_{\mathrm{name}}+r_{\mathrm{para}}+r_{\mathrm{value}}}{1+|T|+\sum_{T_i\in T}|\mathbf{v}(T_i)|}.
\end{equation}

We define the mixed reward
\begin{equation}
R_j^{\mathrm{mix}} = R_j^{\mathrm{out}} + R_j^{\mathrm{reasoning}},
\end{equation}
where $R_j^{\mathrm{reasoning}}\in [0,1]$ is assigned by an LLM-as-a-Judge and evaluates the coherence and completeness of the generated chain-of-thought (Appendix~\ref{judge}).

For each prompt group $g\in\{1,\dots,G\}$, we sample $K$ responses $\{y_{g,1},\dots,y_{g,K}\}\sim\pi_\theta(\cdot\mid x_g)$.
For $t\in\{\mathrm{out},\mathrm{mix}\}$ with per-sample reward $R^t_{g,j}$, we compute the group mean and variance
\begin{equation}
\bar R_g^{\,t} \coloneqq \frac{1}{K}\sum_{j=1}^K R^t_{g,j},
\quad
\widehat V_g^{\,t} \coloneqq \frac{1}{K}\sum_{j=1}^K \bigl(R^t_{g,j}-\bar R_g^{\,t}\bigr)^2.
\end{equation}
We then form normalized within-group advantages
\begin{equation}
A^t_{g,j} \coloneqq \frac{R^t_{g,j}-\bar R_g^{\,t}}{\sqrt{\widehat V_g^{\,t}}+\epsilon}, \qquad t\in\{\mathrm{out},\mathrm{mix}\},
\end{equation}
where $\epsilon>0$ is a numerical stabilization constant.

\paragraph{Variance-aware gating}
Let $\widehat \sigma^{\,t}_g \coloneqq \sqrt{\widehat V^{\,t}_g}$ denote the within-group reward dispersion for $t\in\{\mathrm{out},\mathrm{mix}\}$.
Following Section~3.1, dispersion ratios provide an observable proxy for mixed-to-outcome alignment, and ratio-based gating yields a conservative alignment-proxy floor (Theorem~\ref{thm:safe_mixing}).
We define the variance-aware gating weight
\begin{equation}
\rho_g\;\coloneqq\; \frac{\widehat \sigma^{\mathrm{mix}}_g}{\widehat \sigma^{\mathrm{out}}_g + \widehat \sigma^{\mathrm{mix}}_g + \varepsilon_{\mathrm{std}}},
\end{equation}
where \(\varepsilon_{\mathrm{std}}>0\) is a small numerical constant.

We combine this reliability gate with saturation-based annealing using only outcome statistics.
Let $R^{\max}_{\mathrm{out}}$ denote the peak observed group-mean outcome reward, and cap the mixed contribution by $\varepsilon_{\mathrm{mix}}$:
\begin{equation}
\label{eq:mix_weight}
w^{\mathrm{mix}}_g \;\coloneqq\; 
\mathbf{1}\!\bigl(\bar R^{\mathrm{out}}_g < R^{\max}_{\mathrm{out}}\bigr) \cdot \mathbf{1}\!\bigl(\rho_g < \varepsilon_{\mathrm{mix}}\bigr) \, \rho_g,
\end{equation}
where $\varepsilon_{\mathrm{mix}}$ bounds the acceptable relative dispersion so that noisy or weakly aligned mixed signals do not dominate learning. Conditioned on passing the gate, we use $\rho_g$ to modulate the strength of auxiliary injection, reflecting a signal--stability trade-off, while the indicators ensure that high-dispersion mixed rewards cannot dominate the update.

\paragraph{Difficulty-aware weighting}
Beyond adaptive mixing, we reweight prompt groups by optimization difficulty to avoid over-updating on near-saturated groups and under-emphasizing informative failures, a behavior also observed in \citet{zhan2025exgrpolearningreasonexperience}.
To keep the curriculum stable and independent of judge noise, we define the group weight using only outcome statistics (see details in Appendix~\ref{proof:weight_clip}):
\begin{align}
\label{Difficulty-aware weighting}
d_g &\coloneqq 
\alpha_{\mathrm{base}} + (\alpha_{\mathrm{prio}} - \alpha_{\mathrm{base}}) \cdot 
\mathbf{1}\!\bigl( \tau_{\mathrm{low}} < \bar R^{\mathrm{out}}_g < \tau_{\mathrm{high}} \bigr),
\end{align}
where $\alpha_{\mathrm{prio}} > \alpha_{\mathrm{base}} > 0$ and $\tau_{\mathrm{low}}, \tau_{\mathrm{high}}$ define the intermediate-difficulty regime.

Combining Eq.~\ref{eq:mix_weight} and Eq.~\ref{Difficulty-aware weighting}, we obtain the final weighted advantage:
\begin{equation}
\label{eq:hyper-adv-final}
A^{\mathrm{hyper}}_{g,j} \;\coloneqq\; d_g \bigl[ (1 - w^{\mathrm{mix}}_g) A^{\mathrm{out}}_{g,j} + w^{\mathrm{mix}}_g A^{\mathrm{mix}}_{g,j} \bigr].
\end{equation}

\paragraph{Dynamic clipping} To stabilize updates as auxiliary reliance increases, we shrink the clip radius based on the minibatch-average mixing weight.
For a minibatch $\mathcal{B}$, let $\bar w_{\mathcal{B}} \coloneqq \frac{1}{|\mathcal{B}|}\sum_{g\in\mathcal{B}} w^{\mathrm{mix}}_g$ and set
\begin{equation}
\label{eq:groupwise-clip-final}
\varepsilon \coloneqq \varepsilon_{\min} + (1 - \bar{w}_{\mathcal{B}})(\varepsilon_{\max} - \varepsilon_{\min}).
\end{equation}

With the importance sampling ratio
$
r_{g,j}(\theta) \coloneqq \frac{\pi_\theta(y_{g,j} \mid x_g)}{\pi_{\theta_{\mathrm{old}}}(y_{g,j} \mid x_g)},
$
we get the final AWPO objective:
\begin{align}
J(\theta) 
&\coloneqq \mathbb{E}_{g \sim \mathcal{G}} \, \mathbb{E}_{y_{g,j} \sim \pi_{\theta}(\cdot \mid x_g)} \Bigl[ \min\Bigl( r_{g,j}(\theta) A^{\mathrm{hyper}}_{g,j}, \nonumber\\
&\operatorname{clip}(r_{g,j}(\theta), 1 - \varepsilon, 1 + \varepsilon) A^{\mathrm{hyper}}_{g,j} \Bigr) \Bigr].
\end{align}
Complete training procedure is summarized in Algorithm~\ref{alg:vh-grpo-concise}.

\begin{table}[h]

\caption{Comparison on BFCL benchmark across different methods. Average performance is calculated using the official scripts. The column abbreviations stand for: 
 \textbf{NL} (Non-Live),  \textbf{L} (Live) and \textbf{MT} (Multi-Turn).}
\label{bfcl2}
\centering
\small
\setlength{\tabcolsep}{6pt} 
\begin{tabular}{lcccc}
\toprule
\multirow{2}{*}{\textbf{Method}} & \multirow{2}{*}{\textbf{Overall}} & \multicolumn{2}{c}{\textbf{Single Turn}} & \multirow{2}{*}{\textbf{MT}} \\
\cmidrule(lr){3-4}
& & \textit{NL} & \textit{L} & \\
\midrule
\multicolumn{5}{l}{\textbf{Qwen3-1.7B Models}} \\
\cmidrule(r){1-5} 
Base & {54.70\%} & {83.27\%} & {72.95\%} & {8.62\%} \\
SFT & {57.07\%} & {82.90\%} & {74.06\%} & {14.25\%} \\
SFT+GRPO & {57.19\%} & {85.31\%} & {72.10\%} & {14.88\%} \\
ToolRL & {56.78\%} & {85.52\%} & {71.35\%} & {15.50\%} \\
Dr.GRPO & 57.55\% & 84.06\% & 73.66\% & 15.25\% \\
DAPO & 57.53\% & 84.52\% & 71.92\% & 17.12\% \\
AWPO & \textbf{57.61\%} & {85.52\%} & {70.28\%} & \textbf{18.62\%} \\
\midrule 
\multicolumn{5}{l}{\textbf{Qwen3-8B Models}} \\
\cmidrule(r){1-5}
Base & {66.34\%} & {88.81\%} & {78.54\%} & {33.00\%} \\
SFT & {61.39\%} & {82.58\%} & {73.70\%} & {28.00\%} \\ 
SFT+GRPO & {64.40\%} & {85.90\%} & {75.70\%} & {32.38\%} \\
ToolRL & {68.22\%} & {88.50\%} & {80.85\%} & {36.00\%} \\
Dr.GRPO & {68.15\%} & {88.00\%} & {79.25\%} & {38.12\%} \\ 
DAPO & 67.41\% & 88.27\% & 79.21\% & 35.75\% \\
AWPO & \textbf{69.37\%} & {89.73\%} & {79.16\%} & \textbf{40.50\%} \\
\midrule 
\multicolumn{5}{l}{\textbf{Qwen3-4B-2507 Models}} \\
\cmidrule(r){1-5}
Base & {71.63\%} & {87.98\%} & {79.52\%} & {48.00\%} \\
SFT & {71.99\%} & {87.31\%} & {80.63\%} & {48.25\%} \\
SFT+GRPO & {69.74\%} & {89.46\%} & {80.36\%} & {40.38\%} \\
ToolRL & {70.40\%} & {89.40\%} & {80.63\%} & {41.62\%} \\
Dr.GRPO & {71.97\%} & {87.62\%} & {80.19\%} & {48.62\%} \\
DAPO & 71.15\% & 89.33\% & 79.16\% & 46.25\% \\
AWPO & \textbf{73.20\%} & {87.90\%} & {80.32\%} & \textbf{52.12\%} \\
\bottomrule
\end{tabular}
   \vspace{-1.0em}
\end{table}
\begin{figure*}[ht]
    \centering
    \includegraphics[width=1.00\linewidth]{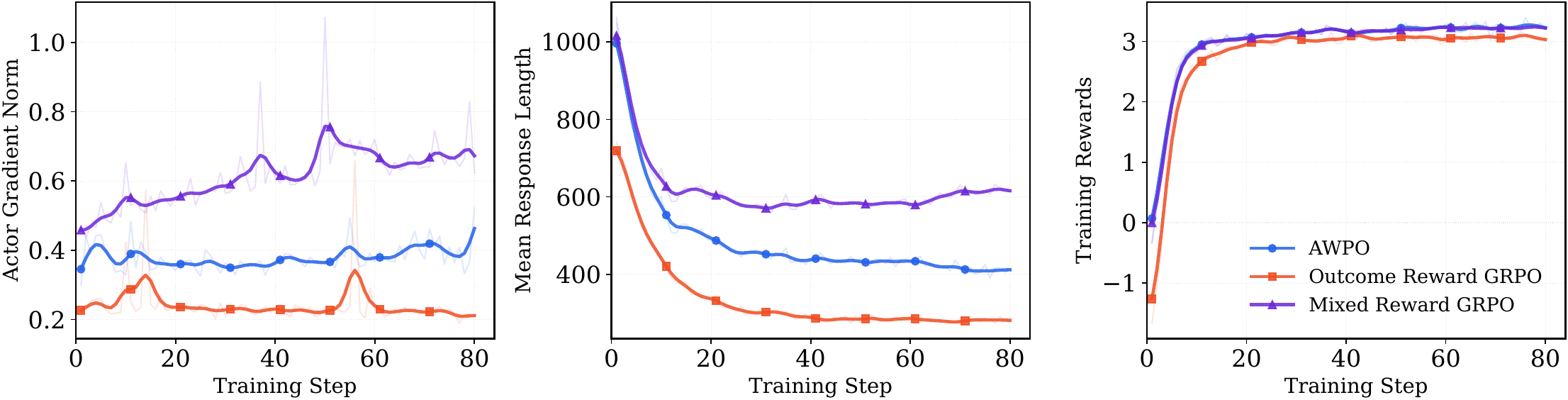}
    \caption{Training dynamics across optimization methods: actor gradient norm, mean response length, and mean reward.}
    \label{fig:3}
    \vspace{-1.5em}
\end{figure*}
\section{Experiments}
\subsection{Experiment Setup}
\paragraph{Training Details} We construct a mixed corpus tailored to robust tool learning in RL. ToolACE \citep{liu2025toolace} focuses on the decision of whether to invoke a tool or answer directly in multi-step interactions; Hammer (masked) \citep{lin2025robusthammer} randomizes tool and argument identifiers to enforce description-grounded generalization rather than lexical memorization; and XLAM \citep{zhang2024xlamfamilylargeaction} provides compositional tasks that require issuing one or multiple tool calls per turn. Jointly, these datasets supervise tool-invocation timing, robust semantic grounding, and compositional planning \citep{qian2025toolrl}.

To obtain denser and more fine-grained reward signals, we convert multi-turn dialogue trajectories into single-step decision instances. Concretely, a dialogue with $K$ interaction steps is decomposed into $K$ sub-instances, each conditioned on the full preceding context as input and supervised with the model’s action at that step as the target. Furthermore, we augment the chain-of-thought component of each sub-instance by calling GLM-4.6 \citep{5team2025glm45agenticreasoningcoding} and GPT-4o \citep{openai2024gpt4ocard} to generate reference rationales, which are subsequently used by an LLM-as-a-Judge module during training. This decomposition strategy, following the SWiRL framework \citep{goldie2025syntheticdatageneration}, substantially increases supervision density: the model receives process-level feedback at every decision point rather than only on the final outcome. Such step-wise supervision leads to more stable learning of multi-step reasoning and tool use. We use Swift \citep{zhao2025swiftascalablelightweightinfrastructure} and Verl \citep{Sheng_2025verl} to train models.

\paragraph{Benchmarks}
We evaluate on two tool-calling benchmarks: BFCL \citep{patil2023bfcl}, which reports single-turn and multi-turn tool-use accuracy, and API-Bank \citep{li2023apibankcomprehensivebenchmarktoolaugmented}, a three-level benchmark of tool invocation in multi-turn dialogues. To isolate algorithmic effects, we fine-tune open-source backbones under matched tool configurations and compare against strong post-training baselines.

To verify that tool-use RL does not compromise general capability, we additionally evaluate on MMLU-Pro \citep{wang2024mmluprorobustchallengingmultitask}, an out-of-distribution multiple-choice benchmark without tool calls, measuring robustness in general language understanding and reasoning.

\paragraph{Baselines}
We compare against supervised and RL post-training baselines. \textbf{Base Model} is the instruction-tuned backbone without task-specific training (zero-shot). \textbf{SFT} applies supervised fine-tuning on the task data with token-level cross-entropy. \textbf{ToolRL(GRPO)} \citep{qian2025toolrl} optimizes a verifiable reward using group-relative, sequence-level advantages. \textbf{SFT+GRPO} runs GRPO from the SFT checkpoint to isolate the incremental effect of RL. \textbf{Dr.GRPO} \citep{liu2025drgrpo} replaces standard-deviation normalization with mean-centering (equivalent to an RLOO-style advantage up to rescaling), mitigating bias toward low-variance prompts and length artifacts while retaining the GRPO update. \textbf{DAPO} \citep{yu2025dapo} further extends GRPO-style training with asymmetric clipping and dynamic sampling, and adds token-level optimization and overlong reward shaping to stabilize long-horizon generation. For external reference, we also report results from similarly sized tool-use models (ToolACE2-8B \citep{liu2025toolace}, BitAgent-8B, watt-tool-8B, ToolACE-MT) and larger general-purpose models (GPT-5, DeepSeek-R1 \citep{deepseekai2025deepseekr1incentivizingreasoningcapability}).

\begin{table}[h]
\caption{Comparison on API-Bank benchmark across different methods. The column abbreviations stand for: 
 \textbf{L1} (Level 1 Acc),  \textbf{L2} (Level 2 Acc) and \textbf{L3} (Level 3 Acc).}
\label{apibank}
\centering
\small
\setlength{\tabcolsep}{6pt} 
\begin{tabular}{lcccc}
\toprule
\textbf{Method} & \textbf{Overall} & \textbf{L1} & \textbf{L2} & \textbf{L3} \\
\midrule
\multicolumn{5}{l}{\textbf{Qwen3-1.7B Models}} \\
\cmidrule(r){1-5} 
Base & 47.57\% & 53.38\% & 28.36\% & 39.69\% \\
SFT & 58.96\% & 66.92\% & 52.24\% & 38.17\% \\
SFT+GRPO & 59.46\% & 65.91\% & 58.21\% & 40.46\% \\
ToolRL & 63.65\% & 70.68\% & 61.19\% & 41.22\% \\
Dr.GRPO & 54.77\% & 62.66\% & 38.81\% & 38.93\% \\
DAPO & 63.48\% & 69.67\% & 59.7\% & 46.56\% \\
AWPO & \textbf{65.16\%} & {71.93\%} & {61.19\%} & {46.56\%} \\
\midrule 
\multicolumn{5}{l}{\textbf{Qwen3-8B Models}} \\
\cmidrule(r){1-5}
Base & 63.32\% & 70.68\% & 53.73\% & 45.80\% \\
SFT & 60.64\% & 69.17\% & 56.72\% & 36.64\% \\
SFT+GRPO & 63.82\% & 72.18\% & 52.22\% & 42.75\% \\
ToolRL & 66.50\% & 75.44\% & 64.18\% & 40.46\% \\
Dr.GRPO & 62.48\% & 71.18\% & 49.25\% & 42.75\% \\
DAPO & 62.98\% & 71.18\% & 52.24\% & 43.51\% \\
AWPO & \textbf{67.67\%} & {72.68\%} & {61.19\%} & {55.73\%} \\
\midrule 
\multicolumn{5}{l}{\textbf{Qwen3-4B-2507 Models}} \\
\cmidrule(r){1-5}
Base & 66.33\% & 72.68\% & 64.18\% & 48.09\% \\ 
SFT & 60.30\% & 67.92\% & 59.70\% & 37.40\% \\ 
SFT+GRPO & 68.51\% & 74.44\% & 65.67\% & 51.91\% \\
ToolRL & 65.33\% & 71.93\% & 62.69\% & 46.56\% \\
Dr.GRPO & 65.66\% & 71.68\% & 61.19\% & 49.62\% \\
DAPO & 67.34\% & 73.43\% & 65.67\% & 49.62\% \\
AWPO & \textbf{68.51\%} & 73.93\% & 65.67\% & 53.44\%  \\
\bottomrule
\end{tabular}
 \vspace{-1.5em}
\end{table}
\begin{table*}[h]
\caption{Ablation study of AWPO components on BFCL. \textbf{Difficulty-aware weighting} corresponds to the difficulty-aware weighting scheme defined in Eq.~\ref{Difficulty-aware weighting}; \textbf{w/o Variance-aware gating} (defined in Eq.~\ref{eq:hyper-adv-final}) refers to the baseline that directly employs the mixed reward for advantage estimation; \textbf{Dynamic clipping} replaces the adaptive clipping radius in Eq.~\ref{eq:groupwise-clip-final} with a fixed clipping range.}
\label{tab:bfcl_ablation}
\centering
\small
\setlength{\tabcolsep}{3pt}
\begin{tabular}{lccccccccc}
\toprule
\multirow{2}{*}{\textbf{Models}} &
\multirow{2}{*}{\textbf{Overall Acc}} &
\multicolumn{5}{c}{\textbf{Multi-Turn}} &
\multicolumn{3}{c}{\textbf{Single-Turn}}\\

\cmidrule(lr){3-7} \cmidrule(lr){8-9}
& & Overall & Base & Miss Function & Miss Parameter & Long Context & Non Live & Live & \\
\midrule
\cmidrule(r){1-10}
 Qwen3-1.7B-AWPO & \textbf{57.61} &\textbf{18.62} & 22.00 & 21.50 & 15.00 & 16.00 &  86.17 & 70.28  \\
 w/o difficulty-aware weighting & 57.27 & 15.62 & 21.50 & 15.50 & 12.00 & 13.50 &  83.60 & 73.48  \\
 w/o variance-aware gating           & 56.92 & 15.50 & 16.50  & 20.00  & 17.00  & 8.50  &  83.73 &  72.28     \\
 w/o dynamic clipping     &  57.54 & 15.87  & 19.00  & 15.00  & 17.50  & 12.00  & 83.44  &  73.92  \\
\midrule
\cmidrule(r){1-10}
 Qwen3-8B-AWPO &   \textbf{69.37} &  \textbf{40.50} & 49.50  & 43.50  & 34.00  & 35.00  & 89.73  & 79.16   \\
 w/o difficulty-aware weighting &  68.11 & 38.62  & 44.50  & 42.00  & 33.50  & 34.50  & 89.44  & 78.50   \\
 w/o variance-aware gating             & 68.13  & 38.12  & 45.00  & 41.50  & 32.50  & 33.50  &  88.98 &  78.99  \\
 w/o dynamic clipping  & 68.83  &  39.62 & 49.00  & 45.00  & 30.00  & 34.50  & 88.81  & 79.08      \\
\midrule
\cmidrule(r){1-10}
 Qwen3-4B-2507-AWPO & \textbf{73.20}  & \textbf{52.12}  & 59.00  & 59.00  & 39.00  & 51.50  & 87.90  & 80.32   \\
 w/o difficulty-aware weighting &  70.05 &  44.62 & 50.50  & 51.50  & 31.00  & 45.50  & 86.69  & 79.52   \\
 w/o variance-aware gating        & 71.51   &  48.12 &  56.00 & 54.00  & 33.50  & 49.00  & 86.81  &  80.05     \\
 w/o dynamic clipping     & 70.71  & 47.25  & 58.00  & 53.50  & 32.00  & 45.50  & 85.81  & 79.48   \\
\bottomrule
\end{tabular}
 \vspace{-1.5em}
\end{table*}

\subsection{Main Results}
Experiment results summarize tool-use performance on BFCL, API-Bank and MMLU-Pro.
In particular, AWPO’s gains concentrate on long-horizon and high-difficulty tool use, while keeping single-turn tool calling and general language understanding stable.

\paragraph{BFCL}
AWPO consistently achieves the best multi-turn BFCL accuracy across all model scales as shown in Table \ref{bfcl2}, suggesting that variance-controlled reasoning rewards are most beneficial when credit assignment is sparse and failures compound across turns. The effect is strongest at 4B-2507, where AWPO raises multi-turn accuracy to 52.12\% and translates it into a +2.80-point overall improvement; single-turn performance remains in the same range, with only a modest shift. 

External leaderboard in Table~\ref{bfcl1} results further highlight AWPO’s parameter efficiency. The advantage is broad across multi-turn subsets (e.g., Miss Func 59.0\% vs.\ 31.0\%; Miss Param 39.0\% vs.\ 28.0\% against ToolACE-2-8B), indicating improvements in robust tool selection and argument grounding rather than superficial format alignment.

\paragraph{API-Bank}
API-Bank Level3 (L3) emphasizes compositional, multi-step tool execution. AWPO’s improvements as shown in Table~\ref{apibank} indicates that advantage-weighted reasoning rewards primarily help long-horizon credit assignment and intermediate decision quality rather than merely improving easy single-call formatting. On Qwen3-8B, AWPO substantially lifts L3 accuracy (40.46\% $\rightarrow$ 55.73\%) while translating to a modest overall gain (66.50\% $\rightarrow$ 67.67\%), suggesting that most of the benefit comes from resolving the hardest multi-step failures. At Qwen3-4B-2507 and Qwen3-1.7B, AWPO matches the best overall score (68.51\%) and improves L3 over the strongest alternatives. 

To assess whether tool-use RL compromises general language understanding, we evaluate AWPO on MMLU-Pro in Table~\ref{mmlu-pro}. AWPO exhibits no measurable degradation compared to the base model and yields small but consistent gains, indicating that AWPO’s reasoning-reward integration does not induce a tool-specific shortcut or erode core knowledge.

\subsection{Ablation Study}
We ablate AWPO on BFCL across three Qwen3 backbones (Table~\ref{tab:bfcl_ablation}) to quantify the roles of difficulty-aware weighting, variance-aware gating, and adaptive clipping. 
Removing difficulty-aware weighting causes the largest multi-turn drop, most notably at Qwen3-4B-2507 (52.12\%$\rightarrow$44.62\% multi-turn overall accuracy), with regressions concentrated on hard subsets (e.g., Miss Parameter and Long Context). This supports Eq.~\ref{Difficulty-aware weighting}, which reallocates learning toward hard but learnable groups. Removing variance-aware gating (computing advantages from the mixed reward in Eq.~\ref{eq:hyper-adv-final}) reduces multi-turn accuracy at all scales (4B: 52.12\%$\rightarrow$48.12\%; 8B: 40.50\%$\rightarrow$38.12\%; 1.7B: 18.62\%$\rightarrow$15.50\%), consistent with naive reward mixing increasing update variance and competing with the outcome objective. Replacing adaptive clipping in Eq.~\ref{eq:groupwise-clip-final} with fixed bounds further lowers multi-turn performance, again most at 4B (52.12\%$\rightarrow$47.25\%) and mainly on high-dispersion subsets. This indicates that adaptive clipping stabilizes mixed-reward updates by limiting overly large steps while retaining useful reasoning signal.

Figure~\ref{fig:3} validates AWPO by tracking the actor gradient norm, mean response length, and mean reward against outcome-only and naive mixed-reward GRPO. Consistent with our theoretical analysis (Lemmas 3.1 and 3.4), the gradient norm curves reveals that Naïve Mixed-Reward GRPO suffers from high volatility, indicative of severe auxiliary perturbations ($\Delta_{g,j}$) and gradient misalignment. In contrast, AWPO maintains stable norms comparable to the outcome-only baseline, confirming that variance-aware gating effectively suppresses high-dispersion conflicts. This stability prevents the judge-induced verbosity observed in the naïve approach ($\approx 600$ tokens), as shown in mean response length curve, where AWPO converges to efficient trajectories ($\approx 400$ tokens). Finally, the training rewards curves confirm that all three methods achieve convergence, demonstrating that AWPO effectively integrates reasoning signals without compromising the stability of the optimization process.

\section{Conclusion}
\label{sec:conclusion}

We investigate how to exploit reasoning supervision when post-training tool-use language models with reinforcement learning. We propose advantage-weighted policy optimization (AWPO), a GRPO-style framework that integrates reasoning rewards into group-relative advantage estimation under an explicit signal--variance trade-off. Guided by our policy-improvement analysis, AWPO introduces variance-aware gated mixing to admit reasoning signals only when they are statistically informative, difficulty-aware weighting to concentrate learning on medium-difficulty groups with the largest optimization headroom, and adaptive clipping to preserve stability as reliance on higher-variance signals increases. Across BFCL and API-Bank, AWPO yields consistent gains while maintaining out-of-distribution performance on MMLU-Pro. Collectively, these results suggest a practical recipe for tool-use RL: treat reasoning feedback as a valuable but heteroscedastic signal, and couple its integration with variance-aware control to convert it from a fragile auxiliary reward into a reliable driver of policy improvement.

\nocite{langley00}

\bibliography{example_paper}
\bibliographystyle{icml2026}

\newpage
\appendix
\onecolumn
\section{Proofs}
\label{proof}
\subsection{Proof of Lemma~\ref{lem:adv_decomp}}

\begin{lemma}[Normalized auxiliary perturbation decomposition (restated)]
For each group $g$ and sample $j$,
\begin{equation}
\label{eq:adv_decomp_app}
A^{\mathrm{mix}}_{g,j}
\;=\;
\frac{\widehat\sigma_g^{\mathrm{out}}}{\widehat\sigma_g^{\mathrm{mix}}}\,A^{\mathrm{out}}_{g,j}
\;+\;
\frac{\widetilde R^{\mathrm{reasoning}}_{g,j}}{\widehat\sigma_g^{\mathrm{mix}}+\epsilon}
\;+\;
\Delta_{g,j},
\end{equation}
where $\Delta_{g,j}$ is a residual term induced by the stabilization constants in the denominators and $\Delta_{g,j}\to 0$ as $\epsilon\to 0$.
\end{lemma}

\begin{proof}
We start from the definitions used in the method.
For each group $g$ and reward type $t\in\{\mathrm{out},\mathrm{mix}\}$, define the group mean and dispersion
\begin{equation}
\bar R_g^{\,t} \;\coloneqq\; \frac{1}{K}\sum_{j=1}^K R^t_{g,j},
\qquad
\widehat\sigma_g^{\,t} \;\coloneqq\; \sqrt{\widehat V_g^{\,t}}
\;\;\;\text{with}\;\;\;
\widehat V_g^{\,t}\coloneqq \frac{1}{K}\sum_{j=1}^K \bigl(R^t_{g,j}-\bar R_g^{\,t}\bigr)^2 .
\end{equation}
Let the centered (mean-subtracted) rewards be
\begin{equation}
\widetilde R^t_{g,j} \;\coloneqq\; R^t_{g,j}-\bar R_g^{\,t}.
\end{equation}
The within-group normalized advantage is
\begin{equation}
\label{eq:def_adv_app}
A^t_{g,j}
\;\coloneqq\;
\frac{R^t_{g,j}-\bar R_g^{\,t}}{\widehat\sigma_g^{\,t}+\epsilon}
\;=\;
\frac{\widetilde R^t_{g,j}}{\widehat\sigma_g^{\,t}+\epsilon}.
\end{equation}

\textbf{Step 1: expand the mixed centered reward.}
By construction,
\begin{equation}
R^{\mathrm{mix}}_{g,j} \;=\; R^{\mathrm{out}}_{g,j} + R^{\mathrm{reasoning}}_{g,j}.
\end{equation}
Averaging over $j$ within the same group $g$ yields
\begin{equation}
\bar R_g^{\,\mathrm{mix}}
=
\frac{1}{K}\sum_{j=1}^K R^{\mathrm{mix}}_{g,j}
=
\frac{1}{K}\sum_{j=1}^K \bigl(R^{\mathrm{out}}_{g,j}+R^{\mathrm{reasoning}}_{g,j}\bigr)
=
\bar R_g^{\,\mathrm{out}} + \bar R_g^{\,\mathrm{reasoning}}.
\end{equation}
Therefore the centered mixed reward decomposes \emph{exactly} as
\begin{align}
\widetilde R^{\mathrm{mix}}_{g,j}
&\coloneqq R^{\mathrm{mix}}_{g,j}-\bar R_g^{\,\mathrm{mix}} \nonumber\\
&=
\bigl(R^{\mathrm{out}}_{g,j}+R^{\mathrm{reasoning}}_{g,j}\bigr)
-
\bigl(\bar R_g^{\,\mathrm{out}}+\bar R_g^{\,\mathrm{reasoning}}\bigr) \nonumber\\
&=
\underbrace{\bigl(R^{\mathrm{out}}_{g,j}-\bar R_g^{\,\mathrm{out}}\bigr)}_{=\widetilde R^{\mathrm{out}}_{g,j}}
+
\underbrace{\bigl(R^{\mathrm{reasoning}}_{g,j}-\bar R_g^{\,\mathrm{reasoning}}\bigr)}_{=\widetilde R^{\mathrm{reasoning}}_{g,j}} \nonumber\\
&=
\widetilde R^{\mathrm{out}}_{g,j} + \widetilde R^{\mathrm{reasoning}}_{g,j}.
\label{eq:centered_mix_sum}
\end{align}

\textbf{Step 2: write $A^{\mathrm{mix}}$ and separate terms.}
Using \eqref{eq:def_adv_app} with $t=\mathrm{mix}$ and \eqref{eq:centered_mix_sum},
\begin{align}
A^{\mathrm{mix}}_{g,j}
&=
\frac{\widetilde R^{\mathrm{mix}}_{g,j}}{\widehat\sigma_g^{\,\mathrm{mix}}+\epsilon}
=
\frac{\widetilde R^{\mathrm{out}}_{g,j}+\widetilde R^{\mathrm{reasoning}}_{g,j}}{\widehat\sigma_g^{\,\mathrm{mix}}+\epsilon} \nonumber\\
&=
\frac{\widetilde R^{\mathrm{out}}_{g,j}}{\widehat\sigma_g^{\,\mathrm{mix}}+\epsilon}
+
\frac{\widetilde R^{\mathrm{reasoning}}_{g,j}}{\widehat\sigma_g^{\,\mathrm{mix}}+\epsilon}.
\label{eq:mix_split}
\end{align}
The second term in \eqref{eq:mix_split} already matches the desired ``aux perturbation'' term.

\textbf{Step 3: express the outcome-centered term via $A^{\mathrm{out}}_{g,j}$.}
From \eqref{eq:def_adv_app} with $t=\mathrm{out}$ we have
\begin{equation}
A^{\mathrm{out}}_{g,j}
=
\frac{\widetilde R^{\mathrm{out}}_{g,j}}{\widehat\sigma_g^{\,\mathrm{out}}+\epsilon}
\qquad\Longleftrightarrow\qquad
\widetilde R^{\mathrm{out}}_{g,j}
=
\bigl(\widehat\sigma_g^{\,\mathrm{out}}+\epsilon\bigr)\,A^{\mathrm{out}}_{g,j}.
\label{eq:recover_centered_out}
\end{equation}
Plugging \eqref{eq:recover_centered_out} into the first term of \eqref{eq:mix_split} gives
\begin{align}
\frac{\widetilde R^{\mathrm{out}}_{g,j}}{\widehat\sigma_g^{\,\mathrm{mix}}+\epsilon}
&=
\frac{\bigl(\widehat\sigma_g^{\,\mathrm{out}}+\epsilon\bigr)A^{\mathrm{out}}_{g,j}}{\widehat\sigma_g^{\,\mathrm{mix}}+\epsilon}
=
\frac{\widehat\sigma_g^{\,\mathrm{out}}}{\widehat\sigma_g^{\,\mathrm{mix}}}\,A^{\mathrm{out}}_{g,j}
+
\Biggl[
\frac{\widehat\sigma_g^{\,\mathrm{out}}+\epsilon}{\widehat\sigma_g^{\,\mathrm{mix}}+\epsilon}
-
\frac{\widehat\sigma_g^{\,\mathrm{out}}}{\widehat\sigma_g^{\,\mathrm{mix}}}
\Biggr]A^{\mathrm{out}}_{g,j}.
\label{eq:anchor_plus_delta}
\end{align}
In the last step we added and subtracted the anchor coefficient
\(\frac{\widehat\sigma_g^{\,\mathrm{out}}}{\widehat\sigma_g^{\,\mathrm{mix}}}\).

\textbf{Step 4: collect terms and define $\Delta_{g,j}$.}
Substituting \eqref{eq:anchor_plus_delta} back into \eqref{eq:mix_split} yields
\begin{align}
A^{\mathrm{mix}}_{g,j}
&=
\frac{\widehat\sigma_g^{\,\mathrm{out}}}{\widehat\sigma_g^{\,\mathrm{mix}}}\,A^{\mathrm{out}}_{g,j}
+
\frac{\widetilde R^{\mathrm{reasoning}}_{g,j}}{\widehat\sigma_g^{\,\mathrm{mix}}+\epsilon}
+
\underbrace{
\Biggl[
\frac{\widehat\sigma_g^{\,\mathrm{out}}+\epsilon}{\widehat\sigma_g^{\,\mathrm{mix}}+\epsilon}
-
\frac{\widehat\sigma_g^{\,\mathrm{out}}}{\widehat\sigma_g^{\,\mathrm{mix}}}
\Biggr]A^{\mathrm{out}}_{g,j}
}_{\eqqcolon\;\Delta_{g,j}}.
\end{align}
This is exactly \eqref{eq:adv_decomp_app}, proving the decomposition.

\textbf{Step 5: show $\Delta_{g,j}\to 0$ as $\epsilon\to 0$.}
Since $\widehat\sigma_g^{\,\mathrm{out}}>0$ and $\widehat\sigma_g^{\,\mathrm{mix}}>0$ whenever the within-group variances are non-degenerate,
\begin{align}
\Delta_{g,j}
&=
\Biggl[
\frac{\widehat\sigma_g^{\,\mathrm{out}}+\epsilon}{\widehat\sigma_g^{\,\mathrm{mix}}+\epsilon}
-
\frac{\widehat\sigma_g^{\,\mathrm{out}}}{\widehat\sigma_g^{\,\mathrm{mix}}}
\Biggr]A^{\mathrm{out}}_{g,j}
=
\Biggl[
\frac{(\widehat\sigma_g^{\,\mathrm{out}}+\epsilon)\widehat\sigma_g^{\,\mathrm{mix}}-\widehat\sigma_g^{\,\mathrm{out}}(\widehat\sigma_g^{\,\mathrm{mix}}+\epsilon)}
{\widehat\sigma_g^{\,\mathrm{mix}}(\widehat\sigma_g^{\,\mathrm{mix}}+\epsilon)}
\Biggr]A^{\mathrm{out}}_{g,j} \nonumber\\
&=
\Biggl[
\frac{\epsilon\widehat\sigma_g^{\,\mathrm{mix}}-\epsilon\widehat\sigma_g^{\,\mathrm{out}}}
{\widehat\sigma_g^{\,\mathrm{mix}}(\widehat\sigma_g^{\,\mathrm{mix}}+\epsilon)}
\Biggr]A^{\mathrm{out}}_{g,j}
=
\epsilon\,
\frac{\widehat\sigma_g^{\,\mathrm{mix}}-\widehat\sigma_g^{\,\mathrm{out}}}
{\widehat\sigma_g^{\,\mathrm{mix}}(\widehat\sigma_g^{\,\mathrm{mix}}+\epsilon)}
\,A^{\mathrm{out}}_{g,j}.
\label{eq:delta_closed_form}
\end{align}
For fixed $(g,j)$, the prefactor multiplying $\epsilon$ in \eqref{eq:delta_closed_form} remains bounded as $\epsilon\to 0$ (assuming $\widehat\sigma_g^{\,\mathrm{mix}}>0$),
hence $\Delta_{g,j}\to 0$ linearly in $\epsilon$.
This concludes the proof.
\end{proof}
\clearpage
\subsection{Proof of Lemma~\ref{lem:corr_ratio}}

\begin{lemma}[Within-group advantage correlation equals a dispersion ratio (restated)]
\label{lem:corr_ratio_app}
Under Assumption~\ref{ass:cov_nonneg} and ignoring $\Delta_{g,j}$ (or taking $\epsilon$ sufficiently small),
the \emph{expected} within-group correlation between normalized advantages satisfies
\begin{equation}
\label{eq:corr_ratio_app}
\mathbb{E}\Big[\mathbb{E}_j\!\left[A^{\mathrm{out}}_{g,j}A^{\mathrm{mix}}_{g,j}\right]\Big]
\;\approx\;
\mathbb{E}\Big[\frac{\widehat\sigma_g^{\mathrm{out}}}{\widehat\sigma_g^{\mathrm{mix}}}\Big]
\;\approx\;
\frac{\sigma_g^{\mathrm{out}}}{\sigma_g^{\mathrm{mix}}}
\;\in\; (0,1],
\end{equation}
where $\sigma_g^{t}$ denotes the population dispersion of $R_g^{t}$.
\end{lemma}
We use $\mathbb{E}[\cdot]$ to denote expectation over the randomness of the $K$ samples in group $g$
(and any randomness in the auxiliary reward, e.g., judge noise), while $\mathbb{E}_j[\cdot]$ remains
the empirical average over $j=1,\dots,K$ within the realized group batch.

\begin{proof}
Fix a group $g$ and abbreviate the within-group empirical mean operator by
\begin{equation}
\mathbb{E}_j[\phi_{g,j}] \;\coloneqq\; \frac{1}{K}\sum_{j=1}^K \phi_{g,j}.
\end{equation}
Recall the centered rewards
\begin{equation}
\widetilde R^{t}_{g,j} \;\coloneqq\; R^{t}_{g,j} - \bar R_g^{\,t},\qquad
t\in\{\mathrm{out},\mathrm{mix},\mathrm{reasoning}\},
\end{equation}
and the dispersions
\begin{equation}
\widehat V_g^{\,t} \;\coloneqq\; \mathbb{E}_j[(\widetilde R^t_{g,j})^2],
\qquad
\widehat\sigma_g^{\,t} \;\coloneqq\; \sqrt{\widehat V_g^{\,t}}.
\end{equation}
The within-group normalized advantages are
\begin{equation}
A^{t}_{g,j}
\;=\;
\frac{\widetilde R^{t}_{g,j}}{\widehat\sigma_g^{\,t}+\epsilon},
\qquad t\in\{\mathrm{out},\mathrm{mix}\}.
\label{eq:def_adv_corr_ratio}
\end{equation}
By construction, $R^{\mathrm{mix}}=R^{\mathrm{out}}+R^{\mathrm{reasoning}}$ and hence
\begin{equation}
\widetilde R^{\mathrm{mix}}_{g,j}
= \widetilde R^{\mathrm{out}}_{g,j} + \widetilde R^{\mathrm{reasoning}}_{g,j}.
\label{eq:centered_mix_sum_corr}
\end{equation}

\textbf{Step 1: expand the within-group product.}
Using \eqref{eq:def_adv_corr_ratio},
\begin{align}
\mathbb{E}_j\!\left[A^{\mathrm{out}}_{g,j}A^{\mathrm{mix}}_{g,j}\right]
&=
\mathbb{E}_j\!\left[
\frac{\widetilde R^{\mathrm{out}}_{g,j}}{\widehat\sigma_g^{\mathrm{out}}+\epsilon}
\cdot
\frac{\widetilde R^{\mathrm{mix}}_{g,j}}{\widehat\sigma_g^{\mathrm{mix}}+\epsilon}
\right] \nonumber\\
&=
\frac{1}{(\widehat\sigma_g^{\mathrm{out}}+\epsilon)(\widehat\sigma_g^{\mathrm{mix}}+\epsilon)}
\;\mathbb{E}_j\!\left[
\widetilde R^{\mathrm{out}}_{g,j}\widetilde R^{\mathrm{mix}}_{g,j}
\right].
\label{eq:prod_reduce_num}
\end{align}

\textbf{Step 2: decompose the numerator into variance + covariance.}
Substitute \eqref{eq:centered_mix_sum_corr} into the numerator:
\begin{align}
\mathbb{E}_j\!\left[\widetilde R^{\mathrm{out}}_{g,j}\widetilde R^{\mathrm{mix}}_{g,j}\right]
&=
\mathbb{E}_j\!\left[
\widetilde R^{\mathrm{out}}_{g,j}\bigl(\widetilde R^{\mathrm{out}}_{g,j}+\widetilde R^{\mathrm{reasoning}}_{g,j}\bigr)
\right] \nonumber\\
&=
\underbrace{\mathbb{E}_j\!\left[(\widetilde R^{\mathrm{out}}_{g,j})^2\right]}_{=\widehat V_g^{\,\mathrm{out}}=(\widehat\sigma_g^{\mathrm{out}})^2}
\;+\;
\underbrace{\mathbb{E}_j\!\left[\widetilde R^{\mathrm{out}}_{g,j}\widetilde R^{\mathrm{reasoning}}_{g,j}\right]}_{\eqqcolon\;\widehat{\mathrm{Cov}}_g(\mathrm{out},\mathrm{reasoning})}.
\label{eq:num_var_cov}
\end{align}
Plugging \eqref{eq:num_var_cov} into \eqref{eq:prod_reduce_num} yields the \emph{exact} identity
\begin{align}
\mathbb{E}_j\!\left[A^{\mathrm{out}}_{g,j}A^{\mathrm{mix}}_{g,j}\right]
&=
\frac{(\widehat\sigma_g^{\mathrm{out}})^2}{(\widehat\sigma_g^{\mathrm{out}}+\epsilon)(\widehat\sigma_g^{\mathrm{mix}}+\epsilon)}
\;+\;
\frac{\widehat{\mathrm{Cov}}_g(\mathrm{out},\mathrm{reasoning})}{(\widehat\sigma_g^{\mathrm{out}}+\epsilon)(\widehat\sigma_g^{\mathrm{mix}}+\epsilon)}.
\label{eq:exact_corr_decomp}
\end{align}

\textbf{Step 3: non-negativity in expectation under Assumption~\ref{ass:cov_nonneg}.}
Assumption~\ref{ass:cov_nonneg} implies the \emph{population} covariance is non-negative:
\begin{equation}
\mathrm{Cov}\!\left(R^{\mathrm{out}}_g,\,R^{\mathrm{reasoning}}_g\right)\ge 0,
\end{equation}
where the covariance is taken w.r.t. the underlying data distribution (and any auxiliary randomness).

Recall that
\[
\widehat{\mathrm{Cov}}_g(\mathrm{out},\mathrm{reasoning})
=\mathbb{E}_j\!\left[\widetilde R^{\mathrm{out}}_{g,j}\widetilde R^{\mathrm{reasoning}}_{g,j}\right]
=\frac{1}{K}\sum_{j=1}^K (R^{\mathrm{out}}_{g,j}-\bar R_g^{\,\mathrm{out}})(R^{\mathrm{reasoning}}_{g,j}-\bar R_g^{\,\mathrm{reasoning}}).
\]
A standard calculation shows that for i.i.d. samples,
\begin{equation}
\mathbb{E}\!\left[\widehat{\mathrm{Cov}}_g(\mathrm{out},\mathrm{reasoning})\right]
=
\frac{K-1}{K}\,
\mathrm{Cov}\!\left(R^{\mathrm{out}}_g,\,R^{\mathrm{reasoning}}_g\right)
\;\ge\; 0.
\label{eq:covhat_expectation}
\end{equation}
Taking expectation of \eqref{eq:exact_corr_decomp} and using \eqref{eq:covhat_expectation} yields
\begin{align}
\mathbb{E}\Big[\mathbb{E}_j\!\left[A^{\mathrm{out}}_{g,j}A^{\mathrm{mix}}_{g,j}\right]\Big]
&\ge
\mathbb{E}\!\left[
\frac{(\widehat\sigma_g^{\mathrm{out}})^2}{(\widehat\sigma_g^{\mathrm{out}}+\epsilon)(\widehat\sigma_g^{\mathrm{mix}}+\epsilon)}
\right].
\label{eq:lower_bound_eps_in_expectation}
\end{align}
When $\epsilon$ is small relative to typical values of $\widehat\sigma_g^{\mathrm{out}}$ and $\widehat\sigma_g^{\mathrm{mix}}$,
the right-hand side is well-approximated by $\mathbb{E}[\widehat\sigma_g^{\mathrm{out}}/\widehat\sigma_g^{\mathrm{mix}}]$.

\textbf{Step 4: a bound showing the quantity lies in $[-1,1]$ (and in $[0,1]$ in expectation).}
By Cauchy--Schwarz,
\begin{align}
\left|\mathbb{E}_j\!\left[A^{\mathrm{out}}_{g,j}A^{\mathrm{mix}}_{g,j}\right]\right|
&\le
\sqrt{\mathbb{E}_j[(A^{\mathrm{out}}_{g,j})^2]\;\mathbb{E}_j[(A^{\mathrm{mix}}_{g,j})^2]} \;\le\; 1.
\end{align}
Therefore $\mathbb{E}_j[A^{\mathrm{out}}_{g,j}A^{\mathrm{mix}}_{g,j}] \in [-1,1]$ for any realized batch.
Moreover, combining this with \eqref{eq:lower_bound_eps_in_expectation} shows
$\mathbb{E}\big[\mathbb{E}_j[A^{\mathrm{out}}_{g,j}A^{\mathrm{mix}}_{g,j}]\big]\in[0,1]$.

\textbf{Step 5: isolate the ``dispersion ratio'' as the leading term.}
From the exact decomposition \eqref{eq:exact_corr_decomp}, we can rewrite
\begin{align}
\mathbb{E}_j\!\left[A^{\mathrm{out}}_{g,j}A^{\mathrm{mix}}_{g,j}\right]
&=
\underbrace{\frac{\widehat\sigma_g^{\mathrm{out}}}{\widehat\sigma_g^{\mathrm{mix}}}}_{\text{dispersion ratio}}
\cdot
\underbrace{\frac{\widehat\sigma_g^{\mathrm{mix}}}{\widehat\sigma_g^{\mathrm{mix}}+\epsilon}}_{\approx 1}
\cdot
\underbrace{\frac{\widehat\sigma_g^{\mathrm{out}}}{\widehat\sigma_g^{\mathrm{out}}+\epsilon}}_{\approx 1}
\;+\;
\frac{\widehat{\mathrm{Cov}}_g(\mathrm{out},\mathrm{reasoning})}{(\widehat\sigma_g^{\mathrm{out}}+\epsilon)(\widehat\sigma_g^{\mathrm{mix}}+\epsilon)}.
\label{eq:ratio_plus_covterm}
\end{align}
If either (i) $\widehat{\mathrm{Cov}}_g(\mathrm{out},\mathrm{reasoning})$ is negligible compared to $(\widehat\sigma_g^{\mathrm{out}})^2$
in expectation, or (ii) we conservatively ignore the covariance contribution to obtain a proxy that depends only on dispersions,
then taking expectation in \eqref{eq:ratio_plus_covterm} gives
\begin{equation}
\mathbb{E}\Big[\mathbb{E}_j\!\left[A^{\mathrm{out}}_{g,j}A^{\mathrm{mix}}_{g,j}\right]\Big]
\;\approx\;
\mathbb{E}\Big[\frac{\widehat\sigma_g^{\mathrm{out}}}{\widehat\sigma_g^{\mathrm{mix}}}\Big].
\end{equation}
For large $K$, $\widehat\sigma_g^{t}$ concentrates around $\sigma_g^{t}$, yielding the approximation
$\mathbb{E}[\widehat\sigma_g^{\mathrm{out}}/\widehat\sigma_g^{\mathrm{mix}}]\approx \sigma_g^{\mathrm{out}}/\sigma_g^{\mathrm{mix}}$.

This establishes \eqref{eq:corr_ratio_app}.
\end{proof}
\clearpage

\subsection{Proof of Lemma~\ref{lem:grad_alignment}}
\label{proof:lemma3}

\begin{assumption}[Bounded Spectrum on the Update Subspace]
\label{ass:gram_iso}
Fix a group $g$ and let $Z_g=[z_{g,1},\ldots,z_{g,K}]\in\mathbb{R}^{d\times K}$ stack the score features
$z_{g,j}=\nabla_\theta\log\pi_\theta(a_{g,j}\mid s_{g,j})$.
Define the within-group Gram matrix $G_g:=Z_g^\top Z_g\in\mathbb{R}^{K\times K}$.

Let $a_g,b_g\in\mathbb{R}^K$ be the coefficient vectors used in Lemma~3.4,
and let $\mathcal{S}_g:=\mathrm{span}\{a_g,b_g\}$ be the subspace spanned by the outcome and mixed advantages.
Let $P_g$ denote the orthogonal projector onto $\mathcal{S}_g$.

We assume that the restriction of the Gram matrix to this subspace is spectrally bounded. Specifically, there exist scalars $\mu_g > 0$ and $\delta_g \in [0, 1)$ such that for all $u \in \mathcal{S}_g$:
\begin{equation}
\label{eq:gram_iso_2d_equiv}
\mu_g(1-\delta_g)\|u\|_2^2 \le u^\top G_g u \le \mu_g(1+\delta_g)\|u\|_2^2.
\end{equation}
\end{assumption}

\begin{remark}
While the Fisher Information Matrix in deep neural networks is typically ill-conditioned globally, Assumption~\ref{ass:gram_iso} only restricts the spectrum on the specific 2D plane $\mathcal{S}_g$ governing the current update. Here, $\delta_g$ serves as a proxy for the subspace condition number $\kappa_g \approx \frac{1+\delta_g}{1-\delta_g}$. A larger $\delta_g$ (close to 1) reflects high anisotropy (ill-conditioning), which tightens the slack term in our lower bound, accurately reflecting that gradient alignment is harder to guarantee in "flat" or "sharp" directions.
\end{remark}

\begin{lemma}[Gradient-alignment proxy via within-group correlation (restated)]
\label{lem:grad_alignment_app}
Assume Assumption~\ref{ass:gram_iso} holds, $\|Z(s,a)\|$ is bounded, and the sampling distribution is fixed within the update.
Then, in expectation over the $K$ samples in group $g$,
\begin{equation}
\label{eq:cos_lower_app}
\mathbb{E}\!\left[\cos\!\bigl(g^{\mathrm{out}}_g,\,g^{\mathrm{mix}}_g\bigr)\right]
\;\gtrsim_{\delta_g}\;
\mathbb{E}\!\left[\mathbb{E}_j\!\left[A^{\mathrm{out}}_{g,j}A^{\mathrm{mix}}_{g,j}\right]\right],
\end{equation}
where $\gtrsim_{\delta_g}$ denotes a conservative lower bound up to group-dependent constants and a slack term controlled only by $\delta_g$
(cf. the explicit bound in \eqref{eq:cos_expect_slack_app}).
\end{lemma}

\begin{proof}
\textbf{Notation.}
Fix a group $g$ and write $z_j:=z_{g,j}$.
Let $a_j:=A^{\mathrm{out}}_{g,j}$ and $b_j:=A^{\mathrm{mix}}_{g,j}$, and stack them as
$\mathbf{a},\mathbf{b}\in\mathbb{R}^K$.
Define the (unclipped) group-wise gradients
\begin{equation}
g^{\mathrm{out}}_g:=\mathbb{E}_j[z_j a_j]=\frac{1}{K}Z_g\mathbf{a},
\qquad
g^{\mathrm{mix}}_g:=\mathbb{E}_j[z_j b_j]=\frac{1}{K}Z_g\mathbf{b},
\end{equation}
and the Gram matrix $G_g:=Z_g^\top Z_g$.

\textbf{Step 1: Gram form of cosine similarity (exact).}
Using the matrix representation above,
\begin{align}
\langle g^{\mathrm{out}}_g, g^{\mathrm{mix}}_g\rangle
&=\Big\langle \tfrac{1}{K}Z_g\mathbf{a},\,\tfrac{1}{K}Z_g\mathbf{b}\Big\rangle
=\tfrac{1}{K^2}\mathbf{a}^\top Z_g^\top Z_g \mathbf{b}
=\tfrac{1}{K^2}\mathbf{a}^\top G_g \mathbf{b},\\
\|g^{\mathrm{out}}_g\|^2&=\tfrac{1}{K^2}\mathbf{a}^\top G_g \mathbf{a},
\qquad
\|g^{\mathrm{mix}}_g\|^2=\tfrac{1}{K^2}\mathbf{b}^\top G_g \mathbf{b}.
\end{align}
Therefore,
\begin{equation}
\label{eq:cos_gram_exact_app}
\cos\!\bigl(g^{\mathrm{out}}_g,g^{\mathrm{mix}}_g\bigr)
=
\frac{\mathbf{a}^\top G_g\mathbf{b}}
{\sqrt{\mathbf{a}^\top G_g\mathbf{a}}\;\sqrt{\mathbf{b}^\top G_g\mathbf{b}}}.
\end{equation}

\textbf{Step 2: Control the Gram quadratic forms via subspace spectral bounds.}
Recall $\mathcal{S}_g=\mathrm{span}\{\mathbf{a},\mathbf{b}\}$ and let $P_g$ be the orthogonal projector onto $\mathcal{S}_g$.
Since $\mathbf{a},\mathbf{b}\in\mathcal{S}_g$, we have $P_g\mathbf{a}=\mathbf{a}$ and $P_g\mathbf{b}=\mathbf{b}$, hence
\[
\mathbf{a}^\top G_g \mathbf{b}=\mathbf{a}^\top (P_g G_g P_g)\mathbf{b},\qquad
\mathbf{a}^\top G_g \mathbf{a}=\mathbf{a}^\top (P_g G_g P_g)\mathbf{a},\qquad
\mathbf{b}^\top G_g \mathbf{b}=\mathbf{b}^\top (P_g G_g P_g)\mathbf{b}.
\]
Define the restriction of the Gram matrix to the update subspace by $\widetilde G_g:=P_g G_g P_g$.
By Assumption~\ref{ass:gram_iso}, there exist $\mu_g>0$ and $\delta_g\in[0,1)$ such that
\[
\bigl\|\widetilde G_g-\mu_g P_g\bigr\|_{\mathrm{op}}\le \delta_g\,\mu_g.
\]
Equivalently, we can write
\begin{equation}
\label{eq:gram_subspace_decomp_app}
\widetilde G_g=\mu_g\,(P_g+E_g),\qquad \|E_g\|_{\mathrm{op}}\le \delta_g,
\end{equation}
where $E_g:=\mu_g^{-1}(\widetilde G_g-\mu_g P_g)$ is self-adjoint and supported on $\mathcal{S}_g$.
Since $P_g$ acts as the identity on $\mathcal{S}_g$, for any $\mathbf{x}\in\mathcal{S}_g$,
\begin{equation}
\label{eq:quad_bounds_app2}
(1-\delta_g)\|\mathbf{x}\|_2^2
\le \mathbf{x}^\top(P_g+E_g)\mathbf{x}
\le (1+\delta_g)\|\mathbf{x}\|_2^2.
\end{equation}
In particular, applying \eqref{eq:quad_bounds_app2} to $\mathbf{a}$ and $\mathbf{b}$ gives
\begin{equation}
\label{eq:denom_upper_app2}
\sqrt{\mathbf{a}^\top(P_g+E_g)\mathbf{a}}\le \sqrt{1+\delta_g}\,\|\mathbf{a}\|_2,
\qquad
\sqrt{\mathbf{b}^\top(P_g+E_g)\mathbf{b}}\le \sqrt{1+\delta_g}\,\|\mathbf{b}\|_2.
\end{equation}
For the numerator, using Cauchy--Schwarz and $\|E_g\|_{\mathrm{op}}\le \delta_g$,
\begin{align}
\mathbf{a}^\top(P_g+E_g)\mathbf{b}
&=\mathbf{a}^\top\mathbf{b}+\mathbf{a}^\top E_g\mathbf{b}\notag\\
&\ge \mathbf{a}^\top\mathbf{b}-\|\mathbf{a}\|_2\|E_g\|_{\mathrm{op}}\|\mathbf{b}\|_2
\ge \mathbf{a}^\top\mathbf{b}-\delta_g\|\mathbf{a}\|_2\|\mathbf{b}\|_2.
\label{eq:num_lower_app2}
\end{align}
Substituting \eqref{eq:gram_subspace_decomp_app}--\eqref{eq:num_lower_app2} into \eqref{eq:cos_gram_exact_app} (noting that $\mu_g$ cancels) yields
\begin{equation}
\label{eq:cos_slack_app}
\cos\!\bigl(g^{\mathrm{out}}_g,g^{\mathrm{mix}}_g\bigr)
\ge
\frac{1}{1+\delta_g}\cdot
\frac{\mathbf{a}^\top\mathbf{b}}{\|\mathbf{a}\|_2\|\mathbf{b}\|_2}
\;-\;
\frac{\delta_g}{1+\delta_g}.
\end{equation}

\textbf{Step 3: relate coefficient cosine to within-group mean product.}
Recall
\begin{equation}
\mathbb{E}_j[a_j b_j]=\frac{1}{K}\mathbf{a}^\top\mathbf{b},
\qquad
\|\mathbf{a}\|_2^2=K\,\mathbb{E}_j[a_j^2],
\qquad
\|\mathbf{b}\|_2^2=K\,\mathbb{E}_j[b_j^2].
\end{equation}
By within-group normalization with vanishing stabilization $\epsilon \to 0$ (Eq.~(19) in the main text),
\begin{equation}
\mathbb{E}_j[a_j^2]
=
\frac{(\widehat\sigma_g^{\mathrm{out}})^2}{(\widehat\sigma_g^{\mathrm{out}}+\epsilon)^2}
\le 1,
\qquad
\mathbb{E}_j[b_j^2]\le 1,
\end{equation}
hence $\|\mathbf{a}\|_2\le \sqrt{K}$ and $\|\mathbf{b}\|_2\le \sqrt{K}$, so $\|\mathbf{a}\|_2\|\mathbf{b}\|_2\le K$.
In the regime where the within-group correlation proxy is nonnegative (the case of interest in our subsequent use of Lemma~\ref{lem:corr_ratio}),
we have $\mathbf{a}^\top\mathbf{b}\ge 0$ and therefore
\begin{equation}
\frac{\mathbf{a}^\top\mathbf{b}}{\|\mathbf{a}\|_2\|\mathbf{b}\|_2}
\ge
\frac{\mathbf{a}^\top\mathbf{b}}{K}
=
\mathbb{E}_j[a_j b_j]
=
\mathbb{E}_j\!\left[A^{\mathrm{out}}_{g,j}A^{\mathrm{mix}}_{g,j}\right].
\label{eq:coeffcos_ge_meanprod_app}
\end{equation}
Combining \eqref{eq:cos_slack_app} and \eqref{eq:coeffcos_ge_meanprod_app} yields the pointwise bound
\begin{equation}
\label{eq:cos_to_meanprod_app}
\cos\!\bigl(g^{\mathrm{out}}_g,g^{\mathrm{mix}}_g\bigr)
\ge
\frac{1}{1+\delta_g}\,
\mathbb{E}_j\!\left[A^{\mathrm{out}}_{g,j}A^{\mathrm{mix}}_{g,j}\right]
-
\frac{\delta_g}{1+\delta_g}.
\end{equation}

\textbf{Step 4: take expectation and conclude the $\gtrsim_{\delta_g}$ relation.}
Taking expectation over the $K$ samples in group $g$ gives the explicit conservative bound
\begin{equation}
\label{eq:cos_expect_slack_app}
\mathbb{E}\!\left[\cos\!\bigl(g^{\mathrm{out}}_g,g^{\mathrm{mix}}_g\bigr)\right]
\ge
\frac{1}{1+\delta_g}\,
\mathbb{E}\!\left[\mathbb{E}_j\!\left[A^{\mathrm{out}}_{g,j}A^{\mathrm{mix}}_{g,j}\right]\right]
-
\frac{\delta_g}{1+\delta_g}.
\end{equation}
Equation \eqref{eq:cos_expect_slack_app} is precisely the meaning of \eqref{eq:cos_lower_app} with the shorthand $\gtrsim_{\delta_g}$,
establishing the lemma.
\end{proof}

\clearpage
\subsection{Proof of Theorem~\ref{thm:safe_mixing}}

\begin{theorem}[Safe mixing condition implied by ratio gating (restated)]
\label{thm:safe_mixing_app}
Define the observable ratio
\(
\rho_g \coloneqq \frac{\widehat\sigma^{\mathrm{mix}}_g}{\widehat\sigma^{\mathrm{out}}_g+\widehat\sigma^{\mathrm{mix}}_g+\varepsilon_{\mathrm{std}}}.
\)
Assume $\varepsilon_{\mathrm{std}}$ is negligible.
If $\rho_g\le \varepsilon_{\mathrm{mix}}<1$, then
\begin{equation}
\label{eq:alignment_threshold_app}
\frac{\widehat\sigma_g^{\mathrm{out}}}{\widehat\sigma_g^{\mathrm{mix}}}
\;\ge\;
\frac{1-\varepsilon_{\mathrm{mix}}}{\varepsilon_{\mathrm{mix}}}
\;\eqqcolon\;
\kappa_{\min},
\end{equation}
and consequently,
\begin{equation}
\label{eq:alignment_proxy_floor_app}
\mathbb{E}\!\left[\cos\!\bigl(g^{\mathrm{out}}_g,\,g^{\mathrm{mix}}_g\bigr)\right]
\;\gtrsim_{\delta_g}\;
\mathbb{E}\!\left[\mathbb{E}_j\!\left[A^{\mathrm{out}}_{g,j}A^{\mathrm{mix}}_{g,j}\right]\right]
\;\gtrsim\;
\kappa_{\min},
\end{equation}
where the last relation uses the dispersion proxy induced by Lemma~\ref{lem:corr_ratio} (ignoring $\Delta_{g,j}$ and taking $\epsilon$ small).
\end{theorem}

\begin{proof}
\textbf{Step 1: rewrite the gating condition as an inequality in dispersions.}
By definition,
\begin{equation}
\rho_g
=
\frac{\widehat\sigma^{\mathrm{mix}}_g}
{\widehat\sigma^{\mathrm{out}}_g+\widehat\sigma^{\mathrm{mix}}_g+\varepsilon_{\mathrm{std}}}.
\end{equation}
Assume $\rho_g\le \varepsilon_{\mathrm{mix}}$ with $\varepsilon_{\mathrm{mix}}\in(0,1)$.
Then
\begin{equation}
\widehat\sigma^{\mathrm{mix}}_g
\le
\varepsilon_{\mathrm{mix}}
\bigl(\widehat\sigma^{\mathrm{out}}_g+\widehat\sigma^{\mathrm{mix}}_g+\varepsilon_{\mathrm{std}}\bigr),
\end{equation}
which rearranges to
\begin{equation}
(1-\varepsilon_{\mathrm{mix}})\widehat\sigma^{\mathrm{mix}}_g
\le
\varepsilon_{\mathrm{mix}}\widehat\sigma^{\mathrm{out}}_g + \varepsilon_{\mathrm{mix}}\varepsilon_{\mathrm{std}}.
\label{eq:rg_rearrange_app2}
\end{equation}

\textbf{Step 2: obtain the dispersion-ratio lower bound.}
Dividing \eqref{eq:rg_rearrange_app2} by $\varepsilon_{\mathrm{mix}}\widehat\sigma^{\mathrm{mix}}_g>0$ yields
\begin{equation}
\frac{\widehat\sigma^{\mathrm{out}}_g}{\widehat\sigma^{\mathrm{mix}}_g}
\ge
\frac{1-\varepsilon_{\mathrm{mix}}}{\varepsilon_{\mathrm{mix}}}
-
\frac{\varepsilon_{\mathrm{std}}}{\widehat\sigma^{\mathrm{mix}}_g}.
\label{eq:ratio_lower_exact_app2}
\end{equation}
Under the assumption that $\varepsilon_{\mathrm{std}}$ is negligible (or $\varepsilon_{\mathrm{std}}/\widehat\sigma^{\mathrm{mix}}_g=o(1)$),
\eqref{eq:ratio_lower_exact_app2} reduces to \eqref{eq:alignment_threshold_app}, proving the first claim.

\textbf{Step 3: translate dispersion control into the correlation proxy.}
By Lemma~\ref{lem:corr_ratio}, under Assumption~\ref{ass:cov_nonneg} and ignoring $\Delta_{g,j}$,
\begin{equation}
\mathbb{E}\!\left[\mathbb{E}_j\!\left[A^{\mathrm{out}}_{g,j}A^{\mathrm{mix}}_{g,j}\right]\right]
\ge
\mathbb{E}\!\left[
\frac{(\widehat\sigma_g^{\mathrm{out}})^2}{(\widehat\sigma_g^{\mathrm{out}}+\epsilon)(\widehat\sigma_g^{\mathrm{mix}}+\epsilon)}
\right].
\label{eq:corr_lower_from_lem33_app}
\end{equation}
When $\epsilon$ is small relative to $\widehat\sigma_g^{\mathrm{out}}$ and $\widehat\sigma_g^{\mathrm{mix}}$,
the right-hand side of \eqref{eq:corr_lower_from_lem33_app} is a conservative proxy of $\widehat\sigma_g^{\mathrm{out}}/\widehat\sigma_g^{\mathrm{mix}}$,
so combining with \eqref{eq:alignment_threshold_app} yields
\begin{equation}
\mathbb{E}\!\left[\mathbb{E}_j\!\left[A^{\mathrm{out}}_{g,j}A^{\mathrm{mix}}_{g,j}\right]\right]
\;\gtrsim\;
\kappa_{\min}.
\label{eq:corr_floor_app}
\end{equation}

\textbf{Step 4: lift the correlation proxy to gradient alignment.}
Lemma~\ref{lem:grad_alignment_app} gives
\begin{equation}
\mathbb{E}\!\left[\cos\!\bigl(g^{\mathrm{out}}_g,g^{\mathrm{mix}}_g\bigr)\right]
\;\gtrsim_{\delta_g}\;
\mathbb{E}\!\left[\mathbb{E}_j\!\left[A^{\mathrm{out}}_{g,j}A^{\mathrm{mix}}_{g,j}\right]\right].
\end{equation}
Combining with \eqref{eq:corr_floor_app} establishes \eqref{eq:alignment_proxy_floor_app}.

\textbf{Step 5: interpretation.}
Equation \eqref{eq:alignment_threshold_app} shows that the gate $\rho_g\le \varepsilon_{\mathrm{mix}}$ enforces a uniform lower bound on the dispersion ratio,
which in turn lower-bounds the within-group correlation proxy and (up to $\delta_g$-controlled slack) the alignment between outcome-induced and mixed-induced
policy-gradient directions. In particular, the gate depends only on observable within-group dispersions and thus can be implemented without additional supervision.
\end{proof}

\clearpage

\subsection{Details for difficulty-aware weighting and dynamic clipping}
\label{proof:weight_clip}

\textbf{Difficulty-aware weighting (Eq.~\ref{Difficulty-aware weighting}).}
Recall that the outcome reward is the sum of a discrete format component and a bounded execution component,
\begin{equation}
R^{\mathrm{out}} \;=\; S^{\mathrm{format}} + S^{\mathrm{exec}},
\qquad
S^{\mathrm{format}}\in\{0,1\},\quad S^{\mathrm{exec}}\in[0,1].
\end{equation}
Let $U\coloneqq S^{\mathrm{format}}$ and denote the per-group format-correctness rate by
\begin{equation}
p_g \;\coloneqq\; \mathbb{P}(U=1\mid g)
\;=\;
\mathbb{E}[U\mid g].
\end{equation}
Since $U$ is Bernoulli conditioned on $g$, its conditional variance is
\begin{equation}
\mathrm{Var}(U\mid g)
\;=\;
\mathbb{E}[U^2\mid g]-\mathbb{E}[U\mid g]^2
\;=\;
p_g - p_g^2
\;=\;
p_g(1-p_g),
\end{equation}
which peaks at intermediate $p_g$ and vanishes as $p_g\to 0$ or $p_g\to 1$.
This observation matters because GRPO constructs within-group centered and normalized coefficients, so the effective preference signal contributed by group $g$ depends on within-group dispersion of $R^{\mathrm{out}}$.

To connect dispersion to observable group statistics, expand the conditional variance of $R^{\mathrm{out}}$:
\begin{align}
\mathrm{Var}(R^{\mathrm{out}}\mid g)
&=
\mathrm{Var}(U+S^{\mathrm{exec}}\mid g)
\nonumber\\
&=
\mathrm{Var}(U\mid g)+\mathrm{Var}(S^{\mathrm{exec}}\mid g)
+2\,\mathrm{Cov}(U,S^{\mathrm{exec}}\mid g).
\label{eq:var_decomp_out_app}
\end{align}
When correctness is close to binary, the format term often dominates the dispersion structure, in the sense that $\mathrm{Var}(S^{\mathrm{exec}}\mid g)$ and $|\mathrm{Cov}(U,S^{\mathrm{exec}}\mid g)|$ are typically not large enough to offset the collapse of $\mathrm{Var}(U\mid g)$ near the extremes.
Consequently, if $p_g$ is near $0$ or $1$, most samples in group $g$ share the same format outcome, yielding small within-group dispersion and weak within-group rankings after centering; if $p_g$ is intermediate, groups more frequently contain both correct and incorrect samples, producing sharper within-group comparisons and more actionable gradients.

In practice we do not estimate $p_g$ explicitly, but $\bar R_g^{\mathrm{out}}$ is an outcome-only statistic monotonically related to correctness and is available at no extra cost.
We therefore use $\bar R_g^{\mathrm{out}}$ as a difficulty proxy and prioritize intermediate groups via the piecewise-constant weight
\begin{equation}
d_g
\;\coloneqq\;
\alpha_{\mathrm{base}}
+
(\alpha_{\mathrm{prio}}-\alpha_{\mathrm{base}})
\mathbf{1}\!\bigl(\tau_{\mathrm{low}}<\bar R_g^{\mathrm{out}}<\tau_{\mathrm{high}}\bigr),
\end{equation}
so $d_g\in\{\alpha_{\mathrm{base}},\alpha_{\mathrm{prio}}\}$ with $\alpha_{\mathrm{prio}}>\alpha_{\mathrm{base}}>0$.
Because $d_g$ depends only on outcome statistics, the induced curriculum is decoupled from judge noise.

\textbf{Dynamic clipping (Eq.~\ref{eq:groupwise-clip-final}).}
Let $r_{g,j}(\theta)\coloneqq \pi_\theta(y_{g,j}\mid x_g)/\pi_{\theta_{\mathrm{old}}}(y_{g,j}\mid x_g)$ and consider the GRPO surrogate with coefficient $A^{\mathrm{hyper}}_{g,j}$.
For each sample $(g,j)$, define the scalar factor that multiplies the score function in the gradient as
\begin{equation}
c_{g,j}(\theta)
\;\coloneqq\;
\min\!\Bigl(r_{g,j}(\theta),\operatorname{clip}\!\bigl(r_{g,j}(\theta),1-\varepsilon,1+\varepsilon\bigr)\Bigr)\,A^{\mathrm{hyper}}_{g,j},
\end{equation}
so $|c_{g,j}(\theta)|\le (1+\varepsilon)\,|A^{\mathrm{hyper}}_{g,j}|$.
The per-sample (unclipped) policy-gradient contribution can be written as
\begin{equation}
g_{g,j}(\theta) \;\coloneqq\; Z(s_{g,j},a_{g,j})\,c_{g,j}(\theta),
\qquad
Z(s,a)\coloneqq \nabla_\theta \log \pi_\theta(a\mid s),
\end{equation}
and we denote the minibatch gradient estimator by $\widehat g(\theta)\coloneqq \frac{1}{B}\sum_{(g,j)\in\mathcal{B}} g_{g,j}(\theta)$.

Assume the score is bounded as $\|Z(s,a)\|_2\le G_\infty$ and rewards are bounded so that there exists $C_A$ with $|A^t_{g,j}|\le C_A$ for $t\in\{\mathrm{out},\mathrm{mix}\}$.
Since $w_g^{\mathrm{mix}}\in[0,1]$ and $A^{\mathrm{hyper}}_{g,j}=d_g[(1-w_g^{\mathrm{mix}})A^{\mathrm{out}}_{g,j}+w_g^{\mathrm{mix}}A^{\mathrm{mix}}_{g,j}]$, we have
\begin{align}
|A^{\mathrm{hyper}}_{g,j}|
&=
d_g\bigl|(1-w_g^{\mathrm{mix}})A^{\mathrm{out}}_{g,j}+w_g^{\mathrm{mix}}A^{\mathrm{mix}}_{g,j}\bigr|
\nonumber\\
&\le
d_g\bigl((1-w_g^{\mathrm{mix}})|A^{\mathrm{out}}_{g,j}|+w_g^{\mathrm{mix}}|A^{\mathrm{mix}}_{g,j}|\bigr)
\nonumber\\
&\le
d_g\bigl((1-w_g^{\mathrm{mix}})C_A+w_g^{\mathrm{mix}}C_A\bigr)
=
C_A\,d_g.
\label{eq:Ah_bound_app}
\end{align}
Combining $|c_{g,j}(\theta)|\le (1+\varepsilon)|A^{\mathrm{hyper}}_{g,j}|$ with \eqref{eq:Ah_bound_app} yields
\begin{equation}
|c_{g,j}(\theta)| \;\le\; (1+\varepsilon)\,C_A\,d_g.
\label{eq:c_bound_app}
\end{equation}
Taking norms and using $\|Z\|_2\le G_\infty$ gives a uniform bound on each per-sample gradient contribution:
\begin{equation}
\|g_{g,j}(\theta)\|_2
=
\|Z(s_{g,j},a_{g,j})\|_2\,|c_{g,j}(\theta)|
\;\le\;
G_\infty(1+\varepsilon)\,C_A\,d_g.
\label{eq:gj_bound_app}
\end{equation}
A standard second-moment calculation for the minibatch mean then yields
\begin{align}
\mathbb{E}\bigl[\|\widehat g(\theta)\|_2^2\bigr]
&=
\mathbb{E}\Bigl[\Bigl\|\frac{1}{B}\sum_{(g,j)\in\mathcal{B}} g_{g,j}(\theta)\Bigr\|_2^2\Bigr]
\;\le\;
\frac{1}{B}\,\mathbb{E}\bigl[\|g_{g,j}(\theta)\|_2^2\bigr]
\nonumber\\
&\le
\frac{1}{B}\,\mathbb{E}\bigl[G_\infty^2(1+\varepsilon)^2 C_A^2 d_g^2\bigr]
=
\frac{G_\infty^2}{B}\,(1+\varepsilon)^2\,C_A^2\,\mathbb{E}[d_g^2],
\label{eq:second_moment_bound_app}
\end{align}
and thus $\mathrm{Var}(\widehat g)$ is controlled by an upper bound that scales monotonically with $(1+\varepsilon)^2$.
Since the method increases reliance on mixed rewards through the minibatch-average mixing weight $\bar w_{\mathcal B}$, it reduces $\varepsilon$ as $\bar w_{\mathcal B}$ increases by setting
\begin{equation}
\varepsilon
\;\coloneqq\;
\varepsilon_{\min} + (1 - \bar{w}_{\mathcal B})(\varepsilon_{\max} - \varepsilon_{\min}),
\end{equation}
which makes the factor $(1+\varepsilon)^2$ a decreasing function of $\bar w_{\mathcal B}$ and tightens the conservative second-moment envelope precisely when auxiliary injection is more frequent.

\clearpage
\section{Algorithm Details}
\subsection{Algorithm Design}
\begin{algorithm}[H]
  \caption{AWPO}
  \label{alg:vh-grpo-concise}
\begin{algorithmic}
  \STATE \textbf{Input:} prompt groups $\{(x_g, T_g, c_g^\star)\}_{g=1}^G$, policy $\pi_\theta$, reference policy $\pi_{\theta_{\mathrm{old}}}$, judge $\mathcal{J}$
  \STATE \textbf{Hyper:} $K,\,\epsilon,\,\varepsilon_{\mathrm{std}},\,\varepsilon_{\mathrm{mix}},\,\tau_{\mathrm{low}}<\tau_{\mathrm{high}},\,\alpha_{\mathrm{prio}}>\alpha_{\mathrm{base}},\,\varepsilon_{\min}\le\varepsilon_{\max},\,\eta$
  \STATE \textbf{Opt:} epochs per rollout $E$; (optional) minibatch partition of groups
  \STATE \textbf{State:} $R^{\max}_{\mathrm{out}} \gets -\infty$
  \REPEAT
    \STATE $\theta_{\mathrm{old}} \gets \theta$
    \FOR{$g=1$ \textbf{to} $G$}
      \FOR{$j=1$ \textbf{to} $K$}
        \STATE Sample $y_{g,j} \sim \pi_{\theta_{\mathrm{old}}}(\cdot \mid x_g)$
        \STATE Compute outcome components $(S^{\mathrm{format}}_{g,j}, S^{\mathrm{exec}}_{g,j})$ from $(y_{g,j}, T_g)$
        \STATE $R^{\mathrm{out}}_{g,j} \gets S^{\mathrm{format}}_{g,j} + S^{\mathrm{exec}}_{g,j}$
        \STATE Query judge for reasoning score $R^{\mathrm{reasoning}}_{g,j}\in[0,1]$ using $(x_g, y_{g,j}, c_g^\star, T_g)$
        \STATE $R^{\mathrm{mix}}_{g,j} \gets R^{\mathrm{out}}_{g,j} + R^{\mathrm{reasoning}}_{g,j}$
      \ENDFOR
      \STATE $\bar R_g^{\,\mathrm{out}} \gets \frac{1}{K}\sum_{j=1}^K R^{\mathrm{out}}_{g,j}$;\quad
             $\bar R_g^{\,\mathrm{mix}} \gets \frac{1}{K}\sum_{j=1}^K R^{\mathrm{mix}}_{g,j}$
      \STATE $\widehat V_g^{\,\mathrm{out}} \gets \frac{1}{K}\sum_{j=1}^K (R^{\mathrm{out}}_{g,j}-\bar R_g^{\,\mathrm{out}})^2$;\quad
             $\widehat V_g^{\,\mathrm{mix}} \gets \frac{1}{K}\sum_{j=1}^K (R^{\mathrm{mix}}_{g,j}-\bar R_g^{\,\mathrm{mix}})^2$
      \STATE $\widehat \sigma_g^{\,\mathrm{out}} \gets \sqrt{\widehat V_g^{\,\mathrm{out}}}$;\quad
             $\widehat \sigma_g^{\,\mathrm{mix}} \gets \sqrt{\widehat V_g^{\,\mathrm{mix}}}$
    \ENDFOR
    \STATE $R^{\max}_{\mathrm{out}} \gets \max\!\Bigl(R^{\max}_{\mathrm{out}},\,\max_{g} \bar R_g^{\,\mathrm{out}}\Bigr)$
    \FOR{$g=1$ \textbf{to} $G$}
      \STATE $\rho_g \gets \dfrac{\widehat \sigma^{\mathrm{mix}}_g}{\widehat \sigma^{\mathrm{out}}_g + \widehat \sigma^{\mathrm{mix}}_g + \varepsilon_{\mathrm{std}}}$
      \STATE $w^{\mathrm{mix}}_g \gets
      \mathbf{1}\!\bigl(\bar R^{\mathrm{out}}_g < R^{\max}_{\mathrm{out}}\bigr)\cdot
      \mathbf{1}\!\bigl(\rho_g < \varepsilon_{\mathrm{mix}}\bigr)\, \rho_g$
      \STATE $d_g \gets \alpha_{\mathrm{base}} + (\alpha_{\mathrm{prio}}-\alpha_{\mathrm{base}})\cdot
      \mathbf{1}\!\bigl(\tau_{\mathrm{low}}<\bar R^{\mathrm{out}}_g<\tau_{\mathrm{high}}\bigr)$
    \ENDFOR
    \FOR{$e=1$ \textbf{to} $E$}
      \STATE $\bar w_{\mathcal{B}} \gets \frac{1}{G}\sum_{g=1}^G w^{\mathrm{mix}}_g$
      \STATE $\varepsilon \gets \varepsilon_{\min} + (1-\bar w_{\mathcal{B}})(\varepsilon_{\max}-\varepsilon_{\min})$
      \FOR{$g=1$ \textbf{to} $G$}
        \FOR{$j=1$ \textbf{to} $K$}
\STATE $A^{\mathrm{out}}_{g,j} \gets \dfrac{R^{\mathrm{out}}_{g,j}-\bar R_g^{\,\mathrm{out}}}{\widehat \sigma_g^{\,\mathrm{out}}+\epsilon}, \quad A^{\mathrm{mix}}_{g,j} \gets \dfrac{R^{\mathrm{mix}}_{g,j}-\bar R_g^{\,\mathrm{mix}}}{\widehat \sigma_g^{\,\mathrm{mix}}+\epsilon}$
          \STATE $A^{\mathrm{hyper}}_{g,j} \gets d_g\Bigl[(1-w^{\mathrm{mix}}_g)A^{\mathrm{out}}_{g,j}+w^{\mathrm{mix}}_gA^{\mathrm{mix}}_{g,j}\Bigr]$
          \STATE $r_{g,j}(\theta) \gets \dfrac{\pi_\theta(y_{g,j}\mid x_g)}{\pi_{\theta_{\mathrm{old}}}(y_{g,j}\mid x_g)}$
          \STATE $\mathcal{L}_{g,j} \gets \min\Bigl(r_{g,j}(\theta)A^{\mathrm{hyper}}_{g,j},\
          \operatorname{clip}(r_{g,j}(\theta),1-\varepsilon,1+\varepsilon)A^{\mathrm{hyper}}_{g,j}\Bigr)$
        \ENDFOR
      \ENDFOR
      \STATE $J(\theta) \gets \frac{1}{GK}\sum_{g=1}^G\sum_{j=1}^K \mathcal{L}_{g,j}$
      \STATE $\theta \gets \theta + \eta\nabla_\theta J(\theta)$
    \ENDFOR
  \UNTIL{converged}
\end{algorithmic}
\end{algorithm}

\clearpage
\subsection{LLM as a Judge Design}
\label{judge}
\begin{figure}[h]
    \centering
    \includegraphics[width=1.0\linewidth]{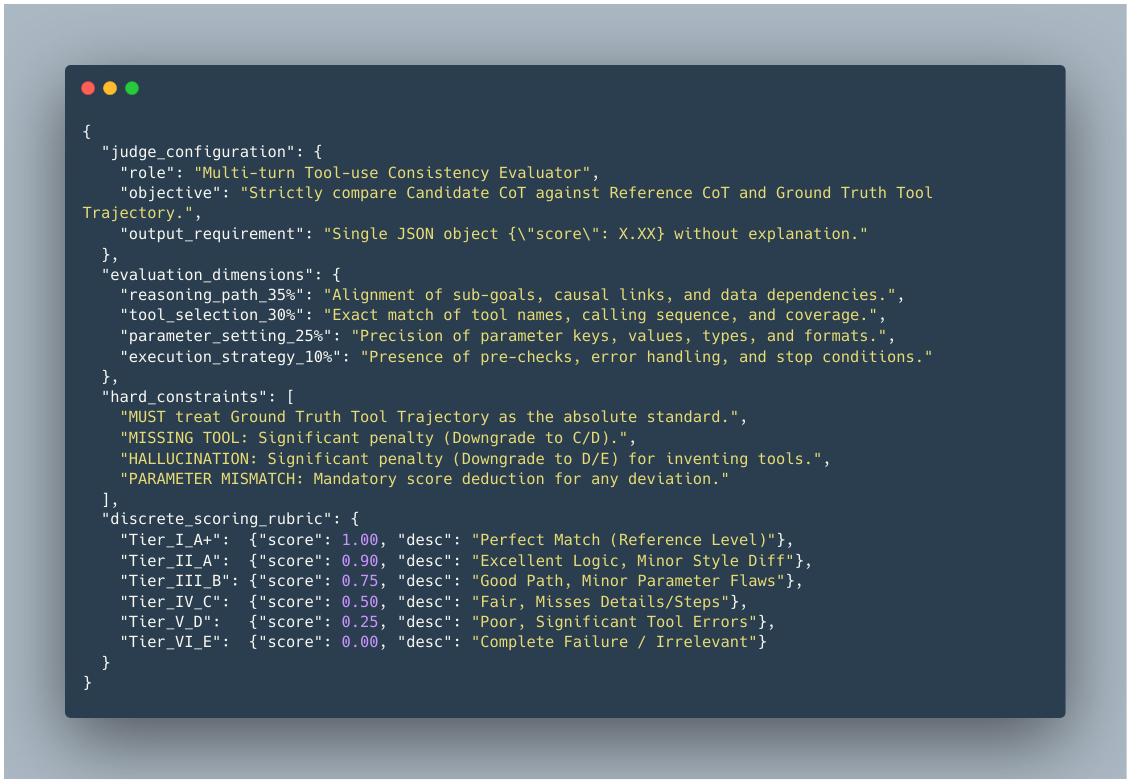}
    \caption{\textbf{Overview of the structured system prompt designed for the Tool-use Consistency Evaluator.} 
    The evaluation protocol is divided into four weighted dimensions: Reasoning Path (35\%), Tool Selection (30\%), Parameter Setting (25\%), and Execution Strategy (10\%). 
    To ensure rigorous assessment, the prompt incorporates \textit{hard constraints} based on ground truth tool trajectories and utilizes a \textit{discrete six-tier scoring rubric} (ranging from Tier VI to Tier I) to map qualitative judgments to fixed numerical values (0.00--1.00), thereby reducing variance in the LLM judge's output.}
    \label{llm judge case}
\end{figure}

We employ the Qwen3-235B-A22B model as the LLM-as-a-Judge to evaluate the quality of the generated chain-of-thought outputs in tool-use tasks. To minimize evaluation errors from the LLM-as-a-Judge as much as possible, we constructed reference chains of thought for the training dataset using GPT-4o. During training, when a model receives a low score, the LLM-as-a-Judge is instructed to specifically assess the degree of deviation between the model’s generated chain of thought and the corresponding reference chain of thought. Moreover, we also draw inspiration from the approach in \citep{li2025evaluatingscoringbiasllmasajudge} that reduces scoring errors of the LLM-as-a-Judge through carefully crafted prompts, and accordingly designed a tailored set of prompts as shown in Figure~\ref{llm judge case}.
\clearpage
\subsection{AWPO Work Flow}
\begin{figure*}[h]
    \centering
    \includegraphics[width=0.99\linewidth]{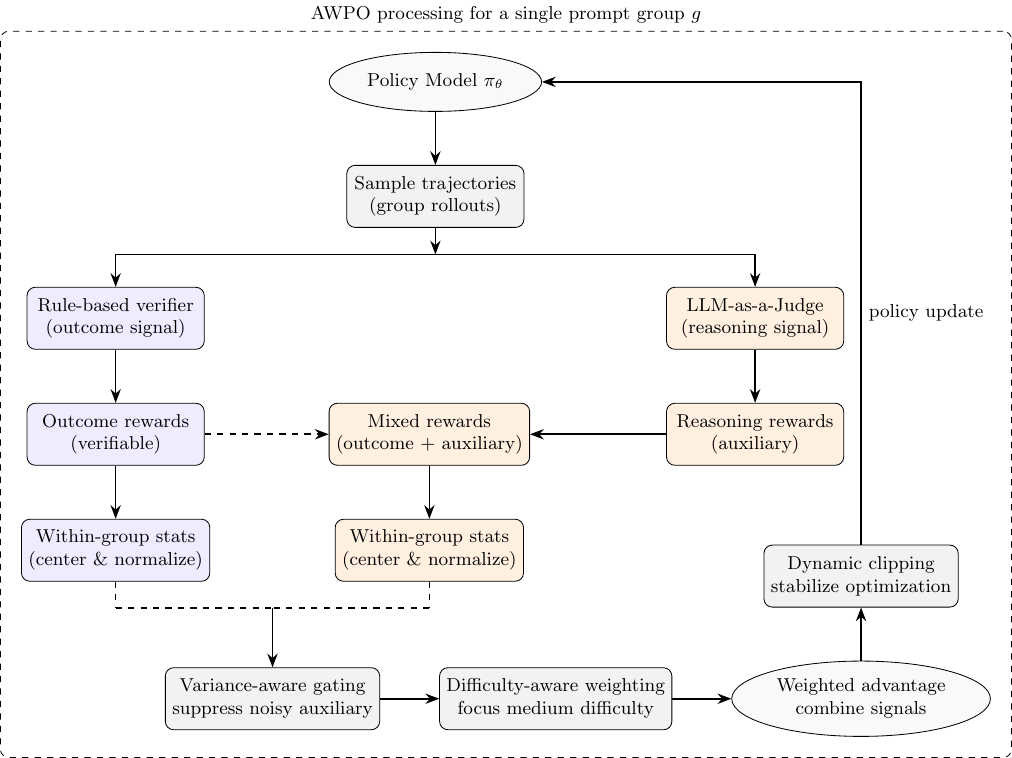}
    \caption{Overview of the AWPO work flow.}
    \label{fig:flow}
\end{figure*}
\clearpage

\section{Experiment Details}
\begin{table*}[h]
    \caption{Comparison on OOD MMLU Pro benchmark. Average performance is calculated using the official scripts.}
\label{mmlu-pro}
\centering
\small
\setlength{\tabcolsep}{6pt} 
\begin{tabular}{lcccccc}
\toprule
\textbf{Models} & \textbf{Overall Accuracy} & \textbf{Math} & \textbf{Physics} & \textbf{Chemistry} & \textbf{Computer Science} & \textbf{Biology} \\
\midrule

Qwen3-1.7B-Base & 48.60\% & 56.77\% & 41.88\%& 43.64\% & 41.71\% & 59.00\%  \\
Qwen3-1.7B-AWPO & \textbf{50.07\%} & {57.81\%} & {43.19\%} & {41.34\%} & 46.34 \%& 61.65 \%  \\
\midrule 
Qwen3-8B-Base & 72.10\% & 80.01\% & 67.90\% & 68.63\% & 65.85\% & 78.10\%   \\
Qwen3-8B-AWPO & \textbf{72.67\%} & {80.46\%} & {68.51\%} & {67.49\%} & {67.80\%} & {79.08\%} \\
\midrule 
Qwen3-4B-2507-Base & 72.37\% & 79.79\% & 70.13\% & 67.05\% & 67.07\% & 77.82\%  \\ 
Qwen3-4B-2507-AWPO & \textbf{73.43\%} & {78.90\%} & {71.29\%} & {68.64\%} & 68.78 \% & 79.50 \%\\
\bottomrule
\end{tabular}
 \vspace{-3pt}
\end{table*}

\begin{table*}[!htbp]
\centering
\caption{Configuration for AWPO training on Qwen3 models.}
\label{Parameter}
\begin{tabular}{ll}
\toprule
\multicolumn{2}{c}{\textbf{Data Configuration}} \\ \midrule
Train Batch Size & 256 \\
Validation Batch Size & 128 \\
Max Prompt Length & 2048 \\
Max Response Length & 2048 \\
\midrule
\multicolumn{2}{c}{\textbf{Optimization}} \\ \midrule
Learning Rate & $1 \times 10^{-6}$ \\
PPO Mini Batch Size & 64 \\
KL Loss Used & False \\
\midrule
\multicolumn{2}{c}{\textbf{Rollout Configuration}} \\ \midrule
Rollout Name & vllm 0.8.5 \\
GPUs & \(8 \times\) 141G H20 or \(8 \times\) 80G A100 \\
GPU Memory Utilization & 0.6 \\
Number of Rollouts & 4 or 8 \\
\midrule
\multicolumn{2}{c}{\textbf{Training \& Logging}} \\ \midrule
Save Frequency (Steps) & 15 \\
Test Frequency (Steps) & 15 \\
Total Epochs & 5 \\
\midrule
\multicolumn{2}{c}{\textbf{Difficulty-aware Weighting Factor}} \\ \midrule
$\alpha_{\mathrm{base}}$: \text{baseline weight scale} & 0.5 \\
$\alpha_{\mathrm{prio}}$: \text{prioritized weight scale} & 1.5 \\
\midrule
\multicolumn{2}{c}{\textbf{Dynamic clipping Configuration}} \\ \midrule
$\varepsilon_{\min}$: \text{minimum clip radius} & 0.18 \\
$\varepsilon_{\max}$: \text{maximum clip radius} & 0.20 \\
\bottomrule
\end{tabular}
\end{table*}

\end{document}